\documentclass[lettersize,journal]{IEEEtran}
\usepackage{amsmath,amsfonts}
\usepackage{array}
\usepackage[caption=false,font=normalsize,labelfont=sf,textfont=sf]{subfig}
\usepackage{textcomp}
\usepackage{stfloats}
\usepackage{url}
\usepackage{verbatim}
\usepackage{graphicx}
\usepackage{colortbl}
\usepackage{booktabs}
\usepackage{multirow}
\usepackage{multicol}
\usepackage{stfloats}
\usepackage{algorithm}
\usepackage{algorithmicx} 
\usepackage{algpseudocode}
\usepackage{cite}

\def\BibTeX{{\rm B\kern-.05em{\sc i\kern-.025em b}\kern-.08em
    T\kern-.1667em\lower.7ex\hbox{E}\kern-.125emX}}
\usepackage{balance}
\begin{document}
\title{DomainForensics: Exposing Face Forgery across Domains via Bi-directional Adaptation}
\author{Qingxuan Lv, Yuezun Li, Junyu Dong, Sheng Chen,~\emph{Life Fellow, IEEE}, Hui Yu, Huiyu Zhou, Shu Zhang
\thanks{{\em Corresponding authors}: Yuezun Li and Junyu Dong} %
\thanks{Q. Lv, Y. Li, J. Dong and S. Zhang are with the College of Computer Science and Technology, Ocean University of China, Qingdao 266100, China (E-mails: lvqingxuan@stu.ouc.edu.cn, liyuezun@ouc.edu.cn, dongjunyu@ouc.edu.cn, zhangshu@ouc.edu.cn).} %
\thanks{S. Chen is with School of Electronics and Computer Science, University of Southampton, Southampton SO17 1BJ, U.K., and also with the College of Computer Science and Technology, Ocean University of China, Qingdao 266100, China (E-mail: sqc@ecs.soton.ac.uk).} %
\thanks{H. Yu is with School of Creative Technologies, Faculty of Creative and Cultural Industries, University of Portsmouth, Portsmouth PO1 2DJ, U.K. (E-mail: hui.yu@port.ac.uk).} %
\thanks{H. Zhou is with School of Computing and Mathematic Sciences, University of Leicester, Leicester LE1 7RH, U.K. (E-mail: hz143@leicester.ac.uk).} %
\vspace*{-5mm}
}

\maketitle

\newcommand{\mb}[1]{\mathbb{#1}}
\newcommand{\mf}[1]{\mathbf{#1}}
\newcommand{\mc}[1]{\mathcal{#1}}
\newcommand{\mr}[1]{\mathrm{#1}}
\newcommand{\mrb}[1]{\mathrm{\mathbf{#1}}}

\def\eg{\emph{e.g.}}
\def\etal{\emph{et al. }}
\def\ie{\emph{i.e.}}

\definecolor{mygray}{gray}{.9}
\definecolor{myyellow}{RGB}{251,255,216}

\begin{abstract}
Recent DeepFake detection methods have shown excellent performance on public datasets but are significantly degraded on new forgeries. Solving this problem is important, as new forgeries emerge daily with the continuously evolving generative techniques. Many efforts have been made for this issue by seeking the commonly existing traces empirically on data level. In this paper, we rethink this problem and propose a new solution from the unsupervised domain adaptation perspective. Our solution, called {\em DomainForensics}, aims to transfer the forgery knowledge from known forgeries (fully labeled source domain) to new forgeries (label-free target domain). Unlike recent efforts, our solution does not focus on data view but on learning strategies of DeepFake detectors to capture the knowledge of new forgeries through the alignment of domain discrepancies. In particular, unlike the general domain adaptation methods which consider the knowledge transfer in the semantic class category, thus having limited application, our approach captures the subtle forgery traces. We describe a new bi-directional adaptation strategy dedicated to capturing the forgery knowledge across domains. Specifically, our strategy considers both forward and backward adaptation, to transfer the forgery knowledge from the source domain to the target domain in forward adaptation and then reverse the adaptation from the target domain to the source domain in backward adaptation. In forward adaptation, we perform supervised training for the DeepFake detector in the source domain and jointly employ adversarial feature adaptation to transfer the ability to detect manipulated faces from known forgeries to new forgeries. In backward adaptation, we further improve the knowledge transfer by coupling adversarial adaptation with self-distillation on new forgeries. This enables the detector to expose new forgery features from unlabeled data and avoid forgetting the known knowledge of known forgery. Extensive experiments demonstrate that our method is surprisingly effective in exposing new forgeries, and can be plug-and-play on other DeepFake detection architectures. 
\end{abstract}


\section{Introduction}\label{S1}

The ever-growing convolutional neural network (CNN) based generative models \cite{goodfellow2014generative,Thies_2016_CVPR,suwajanakorn2017synthesizing,kim2018deep,karras2017progressive,karras2018style} have made face forgery much easier than ever before, allowing people to manipulate the face's identity, appearance and attributes in high realism with little effort. These CNN-based face forgery techniques, known as {\em DeepFake}, have drawn much attention, as their abuse using can lead to impersonation videos, economic fraud, biometric attacks, and even national security problems \cite{survey_chesney_citron_2018}. Thus, it is urgent and important to counteract the misuse of DeepFakes.

\begin{figure}[bp!]
\vspace*{-6mm}
\begin{center}
\includegraphics[width=\linewidth]{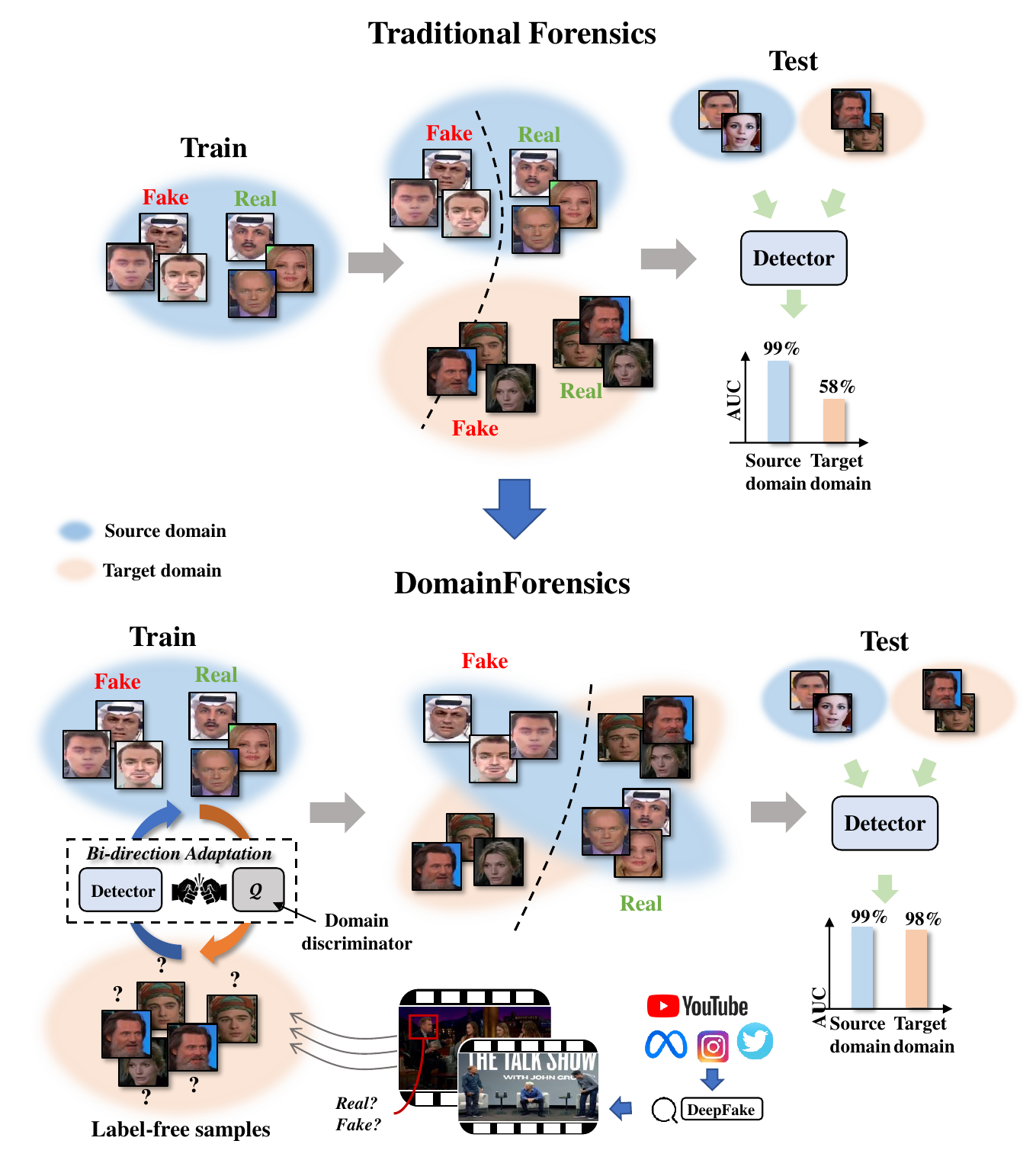}
\end{center}
\vspace{-6mm}
\caption{Overview of traditional forensics (top) and {\em DomainForensics} (bottom). Traditional forensics achieves excellent performance on known forgeries but performs poorly on new forgeries. In contrast, {\em DomainForensics} can effectively expose new forgeries by performing the proposed bi-directional adaption, which can learn the common forgery features across domains using adversarial training.}
\label{fig:framework} 
\vspace*{-1mm}
\end{figure}

During the past few years, large number of DeepFake detection methods \cite{li2018ictu,afchar2018mesonet, xu2023tall, wang2023altfreezing,li2021frequencyaware,sun2021improving,he2021forgerynet} have emerged. Trained on the recently proposed large DeepFake datasets, such as FaceForensics++ (FF++) \cite{rossler2019faceforensics} and Celeb-DF \cite{li2020celeb}, these detection methods have shown promising performance. However, these methods fall into the category that the training and testing sets are from the same distribution, e.g., the same type of forgery or the same dataset, which unfortunately limits their practical applications, as there are always new types of forgeries emerging continuously and widespreading to everywhere on various social platforms. These new types of forgeries are very unlikely to have been included in the existing datasets, and thus they are unseen to these detectors, causing significant performance degradation (see Fig.~\ref{fig:framework} top part). This circumstance gives rise to a big challenge to DeepFake detectors, that is, how to detect constantly emerging new forgeries.

Recently, attempts have been made in the literature to solve this issue. One typical line of research is to use a variety of data augmentation to increase the generalization ability \cite{li2019exposing,Li-etal2020learning,zhao2021learning,shiohara2022detecting}. These methods usually create forged faces by augmenting the pristine videos to cover the known types of forgeries as much as possible. Despite of the promisingly improved generalization, the types of augmentation are limited to known forgeries, thus hindering the performance when confronting unseen forgeries. Frequency clue is also used to improve generalization ability \cite{qian2020thinking,liu2021spatial,luo2021generalizing,dong2022think}. However, this clue is easily affected by data processing and highly correlated with video quality, which cannot perform consistently across different datasets. A different direction of research is to apply transfer learning, such as zero- and few-shot learning \cite{cozzolino2018forensictransfer,aneja2020generalized,qiu2022few}, to improving generalization on new forgeries. Since zero-shot learning cannot access the samples of new forgeries in training, its performance is highly suppressed. In contrast, few-shot learning methods relax the restrictions in that they can access a few samples of new forgeries in training. However, this requires the annotation of these samples, which may not be easily obtained in practice, as we may not know whether a face is forged or not, e.g., multiple faces are in view but only video-level labels are provided. Thus, a fundamental question is: {\em can we detect new forgeries by only accessing target samples without any labels, while achieving competitive performance}?

In this paper, we cast DeepFake detection into a new formulation as an {\em unsupervised domain adaptation} problem, by transferring the knowledge from the source domain to the target domain, without using any annotations of target samples in training. This is very different from the existing strategies and it offers significant advantages over them. Specifically, for DeepFake detection, we can treat the known forgeries as the source domain and new forgeries as the target domain, see Fig.~\ref{fig:framework} bottom part. Our goal is to push the DeepFake detector to learn the common forgery features across different domains by only using label-free interested video collections. It is worth noting that this DeepFake detection problem has a significant discrepancy with the general unsupervised domain adaptation problem, as we aim to learn the common forgery feature from the same category of faces (real or fake), which is more subtle than the semantic features of different categories in the general unsupervised domain adaptation problem (e.g., cat, dog, etc.). To this end, we propose a new unsupervised domain adaptation framework, called {\em DomainForensics}, for DeepFake detection. The key to our {\em DomainForensics} is a novel {\em bi-directional adaptation} strategy. This is very different from the existing DeepFake detection framework which only considers one direction to learn the knowledge supervised by the source domain and transfer it to the target domain. However, since the forgery features are subtle, the one-directional adaptation will inevitably lose a certain amount of knowledge \cite{wang2022rethinking, chen2020big, tifs_distillation}, thus limiting the achievable performance on the target domain. To overcome this problem, we design bi-directional adaptation, which first transfers the knowledge from the source domain to the target domain, referred to as forward adaptation, and then reverses the adaptation from the target domain to the source domain, called backward adaptation. The backward adaptation stage utilizes the results of the forward adaptation stage, further explores the knowledge from the target domain, and transfers it back to the source domain. With the mutual adaptation, DeepFake detector can fully grab the common forgery features across domains. 

To verify our idea, we adopt Vision Transformer (ViT) \cite{dosovitskiy2021an} as our DeepFake detector in the experiment, due to its successful application on vision tasks. Other architectures, such as ResNet \cite{resnet_CVPR}, Xception \cite{chollet2017xception} and EfficientNet \cite{tan2019efficientnet}, can also be used in our framework, and this will also be demonstrated. Since the frequency space can reveal the forgery traces \cite{qian2020thinking,liu2021spatial}, we use color images and corresponding frequency-transformed maps as the input. In the forward adaptation stage, we develop a discriminator that is trained together with the DeepFake detector in an adversarial manner, where the discriminator aims to tell which domain the learned feature is from, and the DeepFake detector aims to extract features that confuse the discriminator. By doing so, the distribution of the target domain is pulled close to the source domain. In the backward adaptation stage, the adaptation is reverted. Since no labels are provided, we employ self-distillation \cite{chen2020big} to further excavate the knowledge of the target domain, and then apply the adversarial training to the distilled model, in order to transfer the knowledge back to the source domain. Extensive experiments are conducted on FF++ and Celeb-DF datasets in several cross-domain scenarios, including different manipulation methods, datasets and types, to demonstrate the effectiveness of our method.

The contribution of this work is summarized as follows.
\begin{enumerate}
\item We propose a new DeepFake detection solution called {\em DomainForensics} to handle continuously emerged new forgeries. Different from recent efforts, our method focuses on pushing the detectors to learn the common forgery features across domains, that is, to transfer the forgery knowledge from known forgeries to unseen forgeries, instead of empirically blending faces on the data level.
\item We propose a new bi-directional adaptation strategy, which first transfers the forgery knowledge from the source domain to the target domain in forward adaptation, and then reverses the adaptation from the target domain to the source domain in backward adaptation. Since the forgery traces are very subtle, we design the backward adaptation stage to further refine the results obtained from the forward adaptation stage with a self-distillation scheme. 
\item Extensive experiments are conducted on FF++ and Celeb-DF datasets with several cross-domain scenarios, including crossing manipulation methods, crossing datasets, and crossing generative types, to demonstrate the effectiveness of our method. We also study the effects of various adaptation settings, various amounts of training samples and different components, to provide thoughtful insights for the following research.    
\end{enumerate}

The remainder of this paper is organized as follows. Section~\ref{sec:relatedwork} reviews the recent works on DeepFake detection and unsupervised domain adaptation. Section~\ref{S3} details our proposed {\em DomainForensics}, including the problem formulation, network framework and bi-directional adaptation. Section~\ref{S4} offers extensive experiments and elaborates on the experimental results. The paper concludes in Section~\ref{S5}.

\section{Related Work}\label{sec:relatedwork} 

In this section, we first present an overview of the existing deepfake detection approaches. We then provide a brief review of unsupervised domain adaptation and discuss the differences between the previous works and our approach.

\subsection{Deepfake Detection}\label{S2.1}

With the advent of large-scale DeepFake datasets, e.g., \cite{li2020celeb,rossler2019faceforensics}, DeepFake detection has made significant progress in recent years, e.g., \cite{li2018ictu,afchar2018mesonet,li2021frequencyaware,sun2021improving,he2021forgerynet, Li-etal2020learning,shiohara2022detecting, xu2023tall, wang2023altfreezing, agarwal2019protecting,sun2021domain,chen2022selfsupervised,TIFSnew1,TIFSnew2}. One challenging problem in this task is how to detect constantly emerging new forgeries. The methods \cite{li2019exposing,Li-etal2020learning,zhao2021learning,shiohara2022detecting,chen2022selfsupervised,TIFSnew3, wang2023dire} enhance generalization ability by exploring elaborate augmentations on pristine videos, with the aim of covering most of the known forgery types. The limitation of these methods is that the augmentation diversity is restricted to known forgeries. Hence, these methods can hardly handle unknown forgeries. Another vein of methods \cite{qian2020thinking,liu2021spatial,luo2021generalizing,li2021frequencyaware,dong2022think,TIFSnew4,TIFSnew5} utilize frequency features to improve generalization ability. However, frequency features can easily be disrupted by post-processing such as compression \cite{zhu2021face}. Inspired by transfer learning, the methods \cite{cozzolino2018forensictransfer,aneja2020generalized,qiu2022few,TIFSnew2} employ zero-shot and few-shot learning to detect new forgeries. Since zero-shot learning cannot access the samples of new forgeries, its performance gain is severely limited. The few-shot learning needs a small portion of samples and corresponding labels of new forgeries. However, although the video-level label is easily obtained, the face-level label is extremely difficult to obtain in practice. 

\subsection{Unsupervised Domain Adaptation}\label{S2.2}

Unsupervised domain adaptation (UDA) aims to address the challenge of transferring knowledge from a source domain to a target domain when labeled data is scarce or completely absent in the target domain. Ben-David \emph{et al.} \cite{ben2006analysis} theoretically revealed that the cross-domain common features serve as latent representations that encapsulate shared and domain-common features across diverse domains. The primary objective is to diminish or eliminate domain-specific variations while retaining domain-agnostic information. The acquisition of cross-domain common representation enhances the model's reliability to domain shifts by prioritizing task-relevant information that transcends domain-specific discrepancies. Consequently, the model achieves improved generalization to unlabeled target domains, even in the scenarios with limited available data.

The existing works for addressing UDA can be classified to two main forms, namely, the discrepancy-based approach and the adversarial approach. Concretely, discrepancy-based methods encourage the model to align the domain discrepancy by minimizing the metrics that can measure the distribution discrepancy between the source and target domains \cite{DeepDomainConfusion_CDA_5, long2015learning, DeepCORAL_ECCV_CDA_4, zhang2019bridging_mdd}. Inspired by the success of generative adversarial network (GAN) \cite{goodfellow2020generative}, recently developed works employed extra adversarial discriminator to align the domain discrepancy, as the feature distributions of source and target domains can be matched by means of confusing the discriminator \cite{DANN_JMLR_ODA_5, tzeng2017adversarial, xu2020adversarial}. In addition, some state-of-the-art methods build up the feature extractor based on modern transformer structure \cite{yang2023tvt, xu2021cdtrans, sun2022safe_ssrt}, which demonstrates that UDA not only helps traditional CNNs to improve the generalization but also is profitable for transformer-based networks. This motivates us to treat the transformer networks as the cornerstone structure and further explore effective UDA methods for face forgery detection.

Note that the general UDA task targets transferring the knowledge of the semantic class category. By contrast, our approach differs from the aforementioned UDA methods in that we aim to explore the subtle forgery features in the face category only. We also find that the existing adaptation schemes, which only consider the adaptation from the source domain to the target domain, is unlikely to perform well on our task. In contrast, our proposed bi-directional adaptation strategy can further explore the knowledge from the unlabeled data in the target domain, as such mutual adaptation coupled with knowledge transfer with self-distillation enables the model to learn common forgery features across known and new forgeries. To the best of our knowledge, Chen and Tan \cite{chen2021featuretransfer} is the first work that attempted to solve Deepfake detection using unsupervised domain adaptation. However, it is a trivial usage of a naive existing solution without improvement, and hence the detection performance is not satisfied. By contrast, our {\em DomainForensics} adopts a meticulously designed strategy, named bi-directional adaptation, which can fully learn the common forgery features across domains and it is validated under several practical cross-domain scenarios.

\section{DomainForensics}\label{S3}

To achieve continuously exposing new forgeries, we formulate DeepFake detection into an unsupervised domain adaptation problem, which transfers the forgery features from known forgeries to new forgeries, without the need of target labeling. Since the forgery features are very subtle, adapting the general UDA to our task is difficult. As such, we propose a new bi-directional adaptation strategy to fully explore the common forgery features across domains. It is worth noting that our {\em DomainForensics} is plug and play, i.e., it can be applied to other DeepFake detection architectures.

We start with the problem formulation in Subsection~\ref{S3.1}, and then discuss the advantages of our {\em DomainForensics} over other architectures in Subsection~\ref{S3.2}, followed by a performance comparison with existing UDA schemes in Subsection~\ref{S3.3}. This naturally motivates us to introduce the new bi-directional adaptation strategy in Subsection~\ref{S3.4} and our network architecture in Subsection~\ref{S3.5}.

\subsection{Problem Formulation}\label{S3.1}

Let the sets of known forgeries and new forgeries corresponding to two different domains be denoted as the source domain $\mathcal{D}_s$ and target domain $\mathcal{D}_t$, respectively. The source domain $\mathcal{D}_s$ is fully annotated as $\mathcal{D}_s = \big\{\big(x_i^s, y_i^s\big)\big\}_{i=1}^{n_s}$, where $\big(x_i^s, y_i^s\big)$ is the $i$-th pair of sample and its corresponding label (i.e., real or not), and $n_s$ is the number of samples. Differently, the target domain $\mathcal{D}_t$ only contains samples without any annotations, which are given by $\mathcal{D}_t = \big\{x_i^t\big\}_{i=1}^{n_t}$, where $n_t$ is the number of samples. Each domain can be divided into a training set and a testing set as $\big\{\mathcal{D}_s^{'}, \mathcal{D}_s^{''}\big\}$ and $\big\{\mathcal{D}_t^{'}, \mathcal{D}_t^{''}\big\}$, respectively. We employ $\mathcal{D}_s^{'}$ and $\mathcal{D}_t^{'}$ in the training phase and perform evaluation on $\mathcal{D}_s^{''}$ and $\mathcal{D}_t^{''}$ in the testing phase. Note that $\mathcal{D}_s^{''}$ and $\mathcal{D}_t^{''}$ are unseen during training. 

Denote a DeepFake detector as $\mathcal{H} = \mathcal{G} \circ \mathcal{F}$, where $\mathcal{G}$ is the classifier and $\mathcal{F}$ is the feature extractor. Given a face image $x$, the output logits of DeepFake detector $\mathcal{H}$ can be defined as $\mathcal{G}(\mathcal{F}(x; \theta_{\mathcal{F}}); \theta_{\mathcal{G}})$, where $\theta_{\mathcal{F}}$ and $\theta_{\mathcal{G}}$ are the parameters of the feature extractor and classifier, respectively.  The challenge under this scenario lies in transferring the forgery knowledge learned from known forgeries $\mathcal{D}_s$ to new forgeries $\mathcal{D}_t$, in terms of the underlying marginal distribution discrepancy, i.e., different manipulated approaches. Our goal is to push the feature extractor to learn the common forgery features across different domains, without the supervision of target labels, i.e., achieving favorable performance on both $\mathcal{D}_s^{''}$ and $\mathcal{D}_t^{''}$. 

\subsection{DomainForensics versus Existing Architectures}\label{S3.2}

Using data augmentation is the most typical solution to improve the generalizability of detection methods, e.g., FWA \cite{li2019exploring}, Face X-ray \cite{Li-etal2020learning}, SBI \cite{shiohara2022detecting}. These methods attempt to synthesize various pseudo-fake faces to cover known forgery as much as possible. By training on these augmented samples, the detectors can learn the common forgery features. In this scenario, the manipulation operations of new forgeries should be known as a prior, in order to synthesize the applicable pseudo-fake faces. However, the technical details of new forgeries may not be accessible in reality, limiting the application of these existing methods. In our scenario, we do not require the technical details of new forgeries. Instead, we can collect a set of videos that contains new forgery faces, and simply extract all faces in a video without knowing the label (real or fake) of faces. Using these samples, we can align the detectors to learn the transferable knowledge from the known forgery to this new forgery. 

This scenario is practical and useful. For example, we can obtain video sets by searching the keywords, e.g., DeepFake, on video platforms. The obtained videos are likely a mix of real and fake faces due to the natural deviation of search engines, i.e., we cannot ensure whether a video that appeared in search results is real or fake. Even though the search results are perfectly matched, i.e., the video-level annotation is correct, the obtained faces can still be mixed, as a video usually contains multiple faces and we cannot know which face is real or fake if only video-level annotation is given. Under this practical circumstance, our method can expose new forgeries with these unlabeled samples. 

\subsection{Comparison with Existing UDA Schemes}\label{S3.3}

\begin{figure}[t]
\centering
\includegraphics[width=\linewidth]{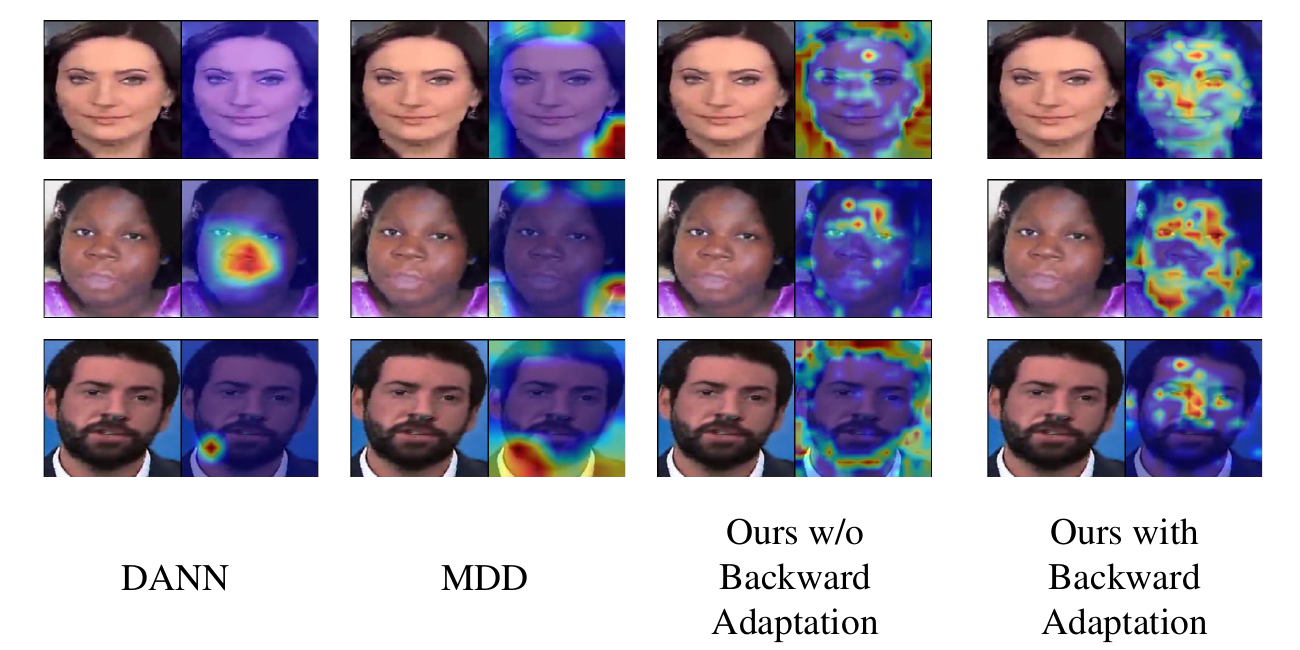}
\vspace{-0.7cm}
\caption{Grad-CAM visualization. We train the models, including DANN \cite{DANN_JMLR_ODA_5}, MDD \cite{zhang2019bridging_mdd} and our {\em DomainForensics}, and visualize the activation maps on FF++ dataset under FS$\rightarrow$F2F scenario. These figures show that models fails to fully capture the common forgery features when only employing one-directional adaptation.}
\label{fig:onedirection} 
\vspace*{-3mm}
\end{figure}

To learn domain-common forgery features, we consider building up a solution from the perspective of UDA. With the aim of reducing the domain discrepancy, early domain adaptation methods put eyes on transferring knowledge from the source domain to the target domain \cite{DANN_JMLR_ODA_5, zhang2019bridging_mdd, DeepCORAL_ECCV_CDA_4}. However, such one-direction adaptation methods are insufficient in digging out the subtle forgery knowledge from unlabeled target data. Fig.~\ref{fig:onedirection} shows several examples of the feature activation maps visualized by Grad-CAM \cite{selvaraju2017grad} using one-direction adaptation. The first two columns from left to right show the CAMs of DANN \cite{DANN_JMLR_ODA_5} and MDD \cite{zhang2019bridging_mdd}, two typical domain adaptation methods. The last two columns are the CAMs of our approach without and with backward adaptation. It can be seen that the typical one-direction domain adaptation cannot locate the forgery features very well in these examples, either paying attention to the background area or the central local face part. In comparison, the CAMs of our method trained with backward adaptation scatter all around the whole face, activating more correct forgery features than these methods including ours without using backward adaptation. This illustration demonstrates that using existing domain adaptation methods for our task is questionable and inspires us to develop a devoted solution, which we detail in the following subsection. 

\subsection{Bi-directional Adaptation}\label{S3.4}

\begin{figure}[t]
\vspace*{-1mm}
\centering
\includegraphics[width=\linewidth]{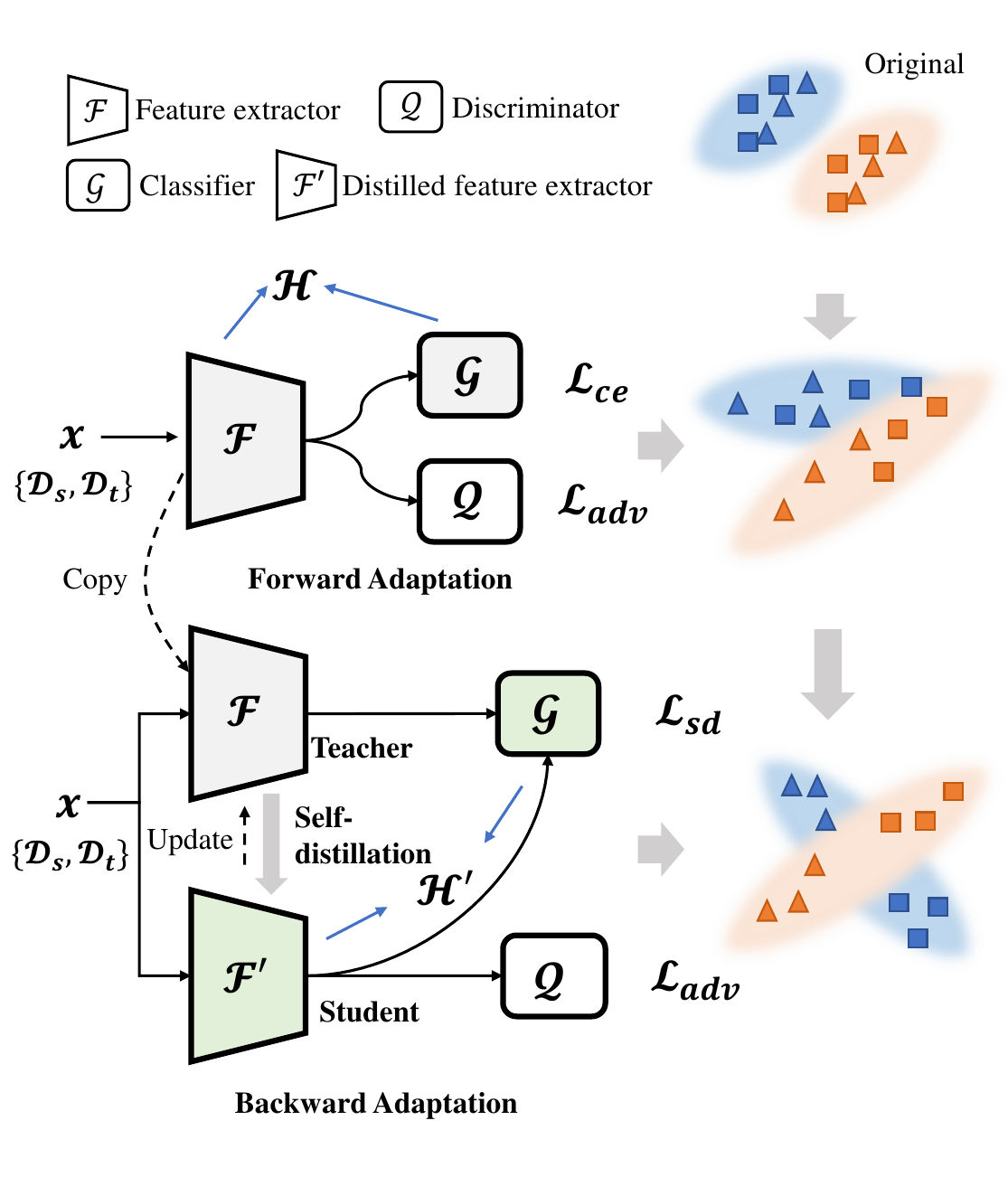}
\vspace{-11mm}
\caption{Illustration of the proposed bi-directional adaptation strategy, containing forward adaptation and backward adaptation with $\mc{H}'$ as the final DeepFake detector. Note that other architectures can also be used in our framework.}
\label{fig:adaptation} 
\vspace*{-2mm}
\end{figure}

As aforementioned, to learn domain-common forgery features, we consider building up a solution from the perspective of UDA. Existing general domain adaptation usually considers one-direction to transfer knowledge from the source domain to the target domain. However, such an one-direction adaptation is insufficient in learning transferable knowledge, as it neglects to learn from unlabeled target data, as demonstrated in the previous subsection. Thus, we propose a new bi-directional adaptation strategy, which consists of both forward adaptation and backward adaptation. The forward adaptation stage aims to transfer the knowledge from the source domain to the target domain, just as the existing solutions \cite{DeepCORAL_ECCV_CDA_4,DANN_JMLR_ODA_5, tzeng2017adversarial}. However, due to the limitation of such an one-direction adaptation, a portion of target domain knowledge is lost in the transfer, thus hindering the performance of the target domain. To eliminate or mitigate this deficiency, we develop a backward adaptation stage to fine-tune the DeepFake detector on the target domain, while retaining the learned knowledge in the forward adaptation stage. Fig.~\ref{fig:adaptation} illustrates the proposed bi-directional adaptation strategy. 

\subsubsection{Forward Adaptation}

In this stage, we aim to learn the common forgery features by adapting the source domain to the target domain. Concretely, we design two loss terms. The first term is a cross-entropy loss $\mathcal{L}_{ce}$, which trains the DeepFake detector on the fully-annotated source domain:
\begin{align}\label{eq1}
  \min\limits_{\theta_{\mathcal{F}},\theta_{\mathcal{G}}} \mathcal{L}_{ce} =& \mathbb{E}_{(x^s_i, y^s_i) \sim \mc{D}_s}\! \left[- \log \mathcal{G}\big(\mathcal{F}(x^s_i; \theta_{\mathcal{F}}); \theta_{\mathcal{G}}\big)^{y^s_i}\right]\! .\!
\end{align}
This loss term enables the DeepFake detector to distinguish classes, i.e., telling apart real and fake faces. 
To transfer this knowledge from the source domain to the target domain, the essence of this stage is to push the feature extractor $\mathcal{F}$ to generate features that cannot be identified from which domain they come. To achieve this goal, we design an adversarial loss $\mathcal{L}_{adv}$ as the second loss term, which guides the training of feature extractor $\mathcal{F}$ with a discriminator $\mathcal{Q}$ in an adversarial manner. Denote $\theta_{\mathcal{Q}}$ as the parameters of discriminator $\mathcal{Q}$. This loss term can be defined as 
\begin{align}\label{eq2}
  & \min\limits_{\theta_{\mathcal{F}}} \max\limits_{\theta_{\mathcal{Q}}} \mathcal{L}_{adv} = \mathbb{E}_{x^s_i \sim \mc{D}_s}\! \left[\log \mathcal{Q}\big(\mathcal{F}(x^s_i; \theta_{\mathcal{F}}); \theta_{\mathcal{Q}}\big)\right] \nonumber \\
  & \hspace*{18mm}+ \mathbb{E}_{x^t_i \sim \mathcal{D}_t}\! \left[\log \big(1 - \mathcal{Q}\big(\mathcal{F}(x^t_i; \theta_{\mathcal{F}}); \theta_{\mathcal{Q}}\big)\big)\right]\! .  
\end{align}
The discriminator $\mathcal{Q}$ outputs binary labels, i.e., $\mathcal{Q}(\cdot; \theta_{\mathcal{Q}}) \in \{0, 1\}$, where $0$ denotes target domain and $1$ denotes source domain. The overall loss of this stage is written as 
\begin{equation}\label{eq3}
  \mathcal{L}_{fas} = \alpha_{1} \cdot \mathcal{L}_{ce} + \alpha_{2} \cdot \mathcal{L}_{adv},
\end{equation}
where $\alpha_1$ and $\alpha_2$ are the weight factors. The training of this stage is an adversarial min-max game between feature extractor $\mathcal{F}$ and discriminator $\mathcal{Q}$. To update $\mathcal{F}$, we fix the parameters of $\mathcal{Q}$, and vice versa. Note that in the optimization of $\mathcal{L}_{adv}$, we do not need any class labels from both domains.

\subsubsection{Backward Adaptation} 

The existing adaptation methods usually only consider the one-direction forward adaptation and their performance are limited on our task due to the loss of knowledge in transferring. The key to overcome this limitation is to mine effective forgery features more specific to the target domain. Concretely, we reverse the adaptation by training the DeepFake detector on the target domain and then transferring the knowledge from the target domain back to the source domain. The major challenge here is that no labels are provided in the target domain, and thus we cannot refine the DeepFake detector in the form of cross-entropy loss $\mathcal{L}_{ce}$ that is used in the forward adaptation stage. 

To address this difficulty, we adopt self-distillation in our framework to further explore the specific representations of the target domain. Inspired by SIMCLRv2 \cite{chen2020big}, we employ a teacher-student network model as the training structure. Specifically, during the training process, we adopt the feature extractor $\mathcal{F}$ with the parameters $\theta_{\mathcal{F}}$ and the classifier $\mathcal{G}$ with the parameters $\theta_{\mathcal{G}}$ used in the forward adaptation stage as the teacher model. We then create a new feature extractor $\mathcal{F}'$ with the parameters $\theta_{\mathcal{F}'}$, which has the same model structure as $\mathcal{F}$ and whose parameters $\theta_{\mathcal{F}'}$ are initialized with the same parameters as $\mathcal{F}$. $\mathcal{F}'$ is combined with the classifier $\mathcal{G}$ to make up the student model. It can be seen that the teacher and student models use the same model structure.

Concretely, the self-distillation loss $\mathcal{L}_{sd}$ is defined as 
\begin{equation}\label{eq4}
  \min\limits_{\theta_{\mathcal{F}},\theta_{\mathcal{F}'},\theta_{\mathcal{G}}}\!\! \mathcal{L}_{sd} = -\!\!\! \sum\limits_{x^t_i \in \mathcal{D}_t}\!\! \left( \sum\limits_{k} \mathcal{P}\big(k|x^t_i;\tau\big) \log \mathcal{P}^{'}\big(k|x^t_i;\tau\big) \right)\! .\!
\end{equation}
Here $\mathcal{P}$ and $\mathcal{P}'$ are the distillation probabilities of the teacher model and student model, respectively. In particular,
\begin{equation}\label{eq5}
  \mathcal{P}(k|x_i;\tau) = \frac{\exp\big(\mathcal{G}(\mathcal{F}(x_i; \theta_{\mathcal{F}}); \theta_{\mathcal{G}})^k / \tau\big)}{\sum_{k'}\exp\big(\mathcal{G}(\mathcal{F}(x_i; \theta_{\mathcal{F}}); \theta_{\mathcal{G}})^{k'} / \tau\big)} ,
\end{equation}
where $k$ is the class index and $\tau$ is a scalar temperature parameter.  $\mathcal{P}'(k|x_i;\tau)$ is obtained by replacing $\mathcal{F}$ with $\mathcal{F}'$ in (\ref{eq5}). 
The Eq.(4) is designed to distill the knowledge from the teacher model learned in the target domain into the student model. By using self-distillation, the student model can enhance the features of the teacher model, thus capturing more effective knowledge regarding the target domain in comparison to the teacher model. We then utilize $\mc{L}_{adv}$ as described in the forward adaptation stage on the same discriminator $\mc{Q}$ and the distilled feature extractor $\mc{F}'$. The overall loss of this stage is the combination of these two loss terms as
\begin{equation}\label{eq6}
 \mc{L}_{bas} = \alpha_3 \cdot \mc{L}_{sd} + \alpha_4 \cdot \mc{L}_{adv}, 
\end{equation}
where $\alpha_3$ and $\alpha_4$ are weight factors. We optimize this loss adversarially as in the forward adaptation stage. At the end of each training epoch, we update the teacher model by copying the parameters of the student model to the teacher model:
\begin{equation}\label{eq7}
  \theta_{\mc{F}} = \theta_{\mc{F}'}.
\end{equation}

This backward adaptation strategy provides more accurate guidance from the teacher model, thus promoting the student model to learn more knowledge. Both the self-distillation loss $\mc{L}_{sd}$ and adversarial loss $\mc{L}_{adv}$ do not need any class labels from both domains.

\subsubsection{Training and Inference} 

The training procedure of our framework is summarized in Algorithm~\ref{alg:biadapt}. In inference, we use feature extractor $\mathcal{F}'$ and classifier $\mathcal{G}$ as our DeepFake detector $\mathcal{H}' = \mathcal{G} \circ \mathcal{F}'$. 

\begin{algorithm}[t]
\small
\caption{Training procedure of bi-directional adaptation}
\label{alg:biadapt} 
\begin{algorithmic}
\Require Source domain $\mathcal{D}_s^{'}$; Target domain $\mathcal{D}_t^{'}$; Initial feature extractor $\mathcal{F}$ with parameters $\theta_{\mathcal{F}}$; Initial classifier $\mathcal{G}$ with parameters $\theta_{\mathcal{G}}$; Initial distilled feature extractor $\mathcal{F}'$ with parameters $\theta_{\mathcal{F}'}$; Initial discriminator $\mathcal{Q}$ with parameters $\theta_{\mathcal{Q}}$; Number of forward adaptation epochs $T_1$; Number of backward adaptation epochs $T_2$
  \State // Forward adaptation stage
  \While{$t \leq T_1$} \Comment{$t = 0$}
    \State // Fix $\mathcal{Q}$
    \State $\min\limits_{\theta_{\mathcal{F}},\theta_{\mathcal{G}}}  \alpha_1 \cdot \mathcal{L}_{ce} + \alpha_1 \cdot \mathcal{L}_{adv}$
    \State // Fix $\mathcal{F}, \mathcal{G}$
    \State $\max\limits_{\theta_{\mathcal{Q}}} \mathcal{L}_{adv}$
  \EndWhile
  \State // Backward adaptation stage
  \While{$t \leq T_2$} \Comment{$t = 0$}
    \State // Fix $\mathcal{Q}$
    \State $\min\limits_{\theta_{\mathcal{F}},\theta_{\mathcal{F}'},\theta_{\mathcal{G}}} \alpha_3 \cdot \mathcal{L}_{sd} + \alpha_4 \cdot \mathcal{L}_{adv}$
    \State // Fix $\mathcal{F}, \mathcal{F}', \mathcal{G}$
    \State $\max\limits_{\theta_{\mathcal{Q}}} \mathcal{L}_{adv}$
    \State // Update $\mathcal{F}$
    \State $\theta_{\mathcal{F}} = \theta_{\mathcal{F}'}$
  \EndWhile
\Ensure $\mathcal{F}', \mathcal{G}$
\end{algorithmic}
\end{algorithm}

\subsection{Network Framework}\label{S3.5}

\begin{figure}[t]
\vspace*{-1mm}
\centering
\includegraphics[width=0.95\linewidth]{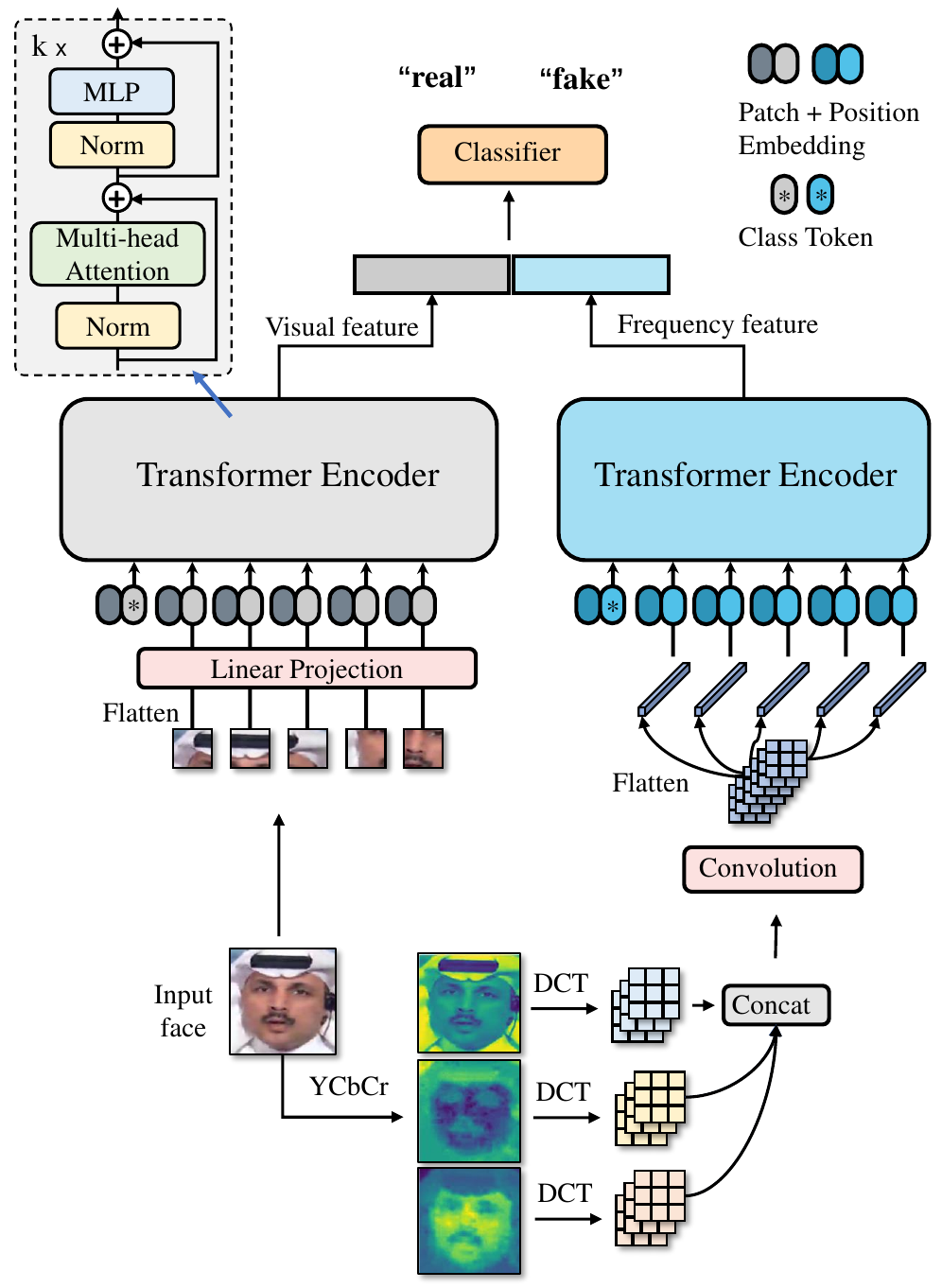}
\vspace{-2mm}
\caption{Network architecture for DeepFake detector.}
\label{fig:network} 
\vspace{-2mm}
\end{figure}

Fig.~\ref{fig:network} depicts the network architecture for our DeepFake detector. We design a ViT-based network as our feature extractor $\mathcal{F}$ due to its strong power on vision tasks. Specifically, our feature extractor has two branches, a visual branch and a frequency branch. For the visual branch, we split the face image into $196$ patches. These patches are flattened and linear transformed to patch embeddings, which are then equipped with position embeddings as the tokens for a ViT \cite{dosovitskiy2021an} architecture to extract visual features. The ViT in this branch contains $l$ transformer layers, each of which is composed by a multi-headed self-attention (MSA) layer and a MLP layer \cite{vaswani2017attention}. For the frequency branch, we first convert the face image from RGB color space to YCbCr color space and then apply DCT transformation to each component with a $8 \times 8$ block \cite{li2021frequencyaware}. The transformed frequency maps are concatenated together and sent into a convolution block. We then flatten the feature maps from the convolution block into $196$ vectors, which are used as the input to another ViT architecture for frequency feature extraction. This ViT architecture contains $m$ transformer layers. The visual features and frequency features are concatenated together as the forgery features for DeepFake detection. The classifier $\mathcal{G}$ has a simple structure with only two linear layers, which takes the forgery features as input and outputs the logit of prediction. It is worth emphasizing that our method is independent of the network architecture, and can be integrated into other mainstream architectures. We use the architecture of Fig.~\ref{fig:network} in our experiments as it can achieve the best performance, which will be confirmed by Table~\ref{tab:backbone_abla} in Ablation Study.

\section{Experiments}\label{S4}

\subsection{Experimental Settings}\label{S4.1}

\subsubsection{Datasets}

We evaluate our approach on three public deepfake detection datasets, which are FF++ \cite{rossler2019faceforensics}, Celeb-DF \cite{li2020celeb}, StyleGAN \cite{stylegan}, DFDCP, and FFIW. 

FF++ is a widely used dataset in deepfake detection. It includes $1,000$ original videos from YouTube, covering a wide range of subjects and scenarios, and consists of four different manipulation techniques: Deepfakes (DF), Face2Face (F2F), FaceSwap (FS), and NeuralTextures (NT), each representing a distinct form of facial manipulation. All these videos have three compression versions, raw, high quality (HQ) and low quality (LQ). From these original videos, $720$ videos, $140$ videos and $140$ videos are used for training, validating and testing, respectively. For each manipulation technique, we employ the same partition as the original videos for training, validating, and testing. We focus on the frame-level deepfake detection and perform evaluation on both HQ and LQ data. As our approach does not rely on extra augmentation operations, we only crop faces from each frame during the preprocessing stage. Concretely, we randomly extract 8 frames for each video clip and crop out the face with 1.3× of the detection box obtained by RetinaFace \cite{Deng2020CVPR}. 

Celeb-DF was proposed more recently, which provides a diverse set of challenges, such as pose, lighting conditions, facial expressions and camera quality, commonly encountered in real-world scenarios. It also offers different levels of manipulation difficulty, ranging from subtle and realistic manipulations to more obvious and noticeable ones. This dataset contains $590$ original videos and $5,639$ DeepFake videos. We respectively use $5,710$ and $518$ videos for training and testing. To construct the training data, we randomly extract $32$ frames from real videos and $4$ frames from fake videos for data balance. As for the testing set, we extract $16$ frames from real and fake videos. The face is cropped out using the same way as for FF++. 

StyleGAN \cite{stylegan} trains a generative adversairal network on Flickr-Faces-HQ dataset (FFHQ), which consists of $70,000$ real face images, to synthesize high-quality fake faces. The synthesizing process focuses on human faces and offers various high-quality synthetic faces by disentangling style and content information in latent space to control the process of face generation. We randomly select $2,500$ original images from FFHQ and $2,500$ synthetic images from the generated dataset, where $2,000$ original images and $2,000$ synthetic images are for training and the remaining images are for testing.

DFDCP \cite{dfdcp_dataset} is a preview dataset for deepfakes detection challenge consisting of $5,000$ videos with two facial modification algorithms, considering diversity in several axes (gender, skin-tone, age, etc.). 

FFIW-10K \cite{zhou2021faceinwild} is a large-scale dataset, which comprises $10,000$ pristine videos and $10,000$ high-quality forgery videos for training. Another $250$ real videos and $250$ manipulated videos are also provided for testing. Different from the aforementioned datasets, this dataset contains multiple forgery faces in a single video.

\subsubsection{Domain Adaptation Scenarios} 

Since a new forgery can either be generated by a new manipulation method or come from an unseen data distribution, we design three adaptation scenarios, which are {\em Cross-manipulation-methods}, {\em Cross-manipulation-datasets} and {\em Cross-manipulation-types}, respectively. Cross-manipulation-methods is the case of crossing different manipulation methods in the same type (e.g., faceswap). Cross-manipulation-datasets represents the adaptation from one dataset to another different one. Cross-manipulation-types is to adapt one type of forgery to a different type. This scenario is more challenging, as different types have significant discrepancies in the manipulation methods, datasets and forged areas, e.g., faceswap faces to GAN-synthesized faces. 

To make a fair comparison with the previous works \cite{Li-etal2020learning,chen2022selfsupervised}, each domain adaptation scenario is trained on the training set and tested on the testing set. Take the Cross-manipulation-methods scenario on FF++ dataset as an example. In training, we use the training set of source and target manipulation methods. In inference, we evaluate the trained model on the testing set of source and target manipulation methods. This configuration also applies to the other two scenarios.

\begin{table}[!t]
\centering
\small
\caption{The performance of our proposed method under two cross-manipulation scenarios on FF++. The top part is the results of O2O and the bottom part is the results of O2M.}
    \begin{tabular}{c|c|c|c|c}
    \toprule
    Adaptation & DF & F2F & NT & FS \\
    \hline \hline
    Baseline & \cellcolor{mygray}100.0 &69.43 & 69.55 & 55.18 \\
    \hline 
    DF $\rightarrow$ F2F & \cellcolor{mygray}99.00 & \cellcolor{myyellow}99.08 & 66.21 & 64.67  \\
    \hline
    DF $\rightarrow$ NT  & \cellcolor{mygray}87.88 & 60.02 & \cellcolor{myyellow}95.03 & 43.49  \\
    \hline
    DF $\rightarrow$ FS  & \cellcolor{mygray}98.82 & 75.68 & 57.41 & \cellcolor{myyellow}99.37  \\
    \hline
    \hline
    Baseline & 81.71 &\cellcolor{mygray}99.13& 55.84 & 69.74 \\
    \hline 
    F2F $\rightarrow$ DF &  \cellcolor{myyellow}99.77 & \cellcolor{mygray}98.67 & 73.56 & 63.55 \\
    \hline
    F2F $\rightarrow$ NT & 89.89 & \cellcolor{mygray}95.59 & \cellcolor{myyellow}95.85 & 57.11  \\
    \hline
    F2F $\rightarrow$ FS & 90.49 & \cellcolor{mygray}98.86 & 51.16 & \cellcolor{myyellow}99.89  \\
    \hline
    \hline
    Baseline & 79.74 & 55.19 & \cellcolor{mygray}95.72 & 44.98 \\
    \hline 
    NT $\rightarrow$ DF  & \cellcolor{myyellow}99.28 & 76.79 &  \cellcolor{mygray}95.80 & 42.94 \\
    \hline
    NT $\rightarrow$ F2F & 92.79 & \cellcolor{myyellow}97.44 & \cellcolor{mygray}93.27 & 63.74  \\
    \hline
    NT $\rightarrow$ FS  & 78.96 & 63.28 & \cellcolor{mygray}70.05 & \cellcolor{myyellow}97.81  \\
    \hline
    \hline
    Baseline & 84.34 & 71.06 & 43.63 & \cellcolor{mygray}99.53\\
    \hline 
    FS $\rightarrow$ DF  & \cellcolor{myyellow}98.90 & 79.62 & 61.16 & \cellcolor{mygray}98.95  \\
    \hline
    FS $\rightarrow$ F2F & 85.40 & \cellcolor{myyellow}99.14 & 50.67 & \cellcolor{mygray}99.69  \\
    \hline
    FS $\rightarrow$ NT  & 88.74 & 73.62 & \cellcolor{myyellow}96.27 & \cellcolor{mygray}95.71  \\
    \hline \hline
    DF $\rightarrow$ F2F,NT,FS & \cellcolor{mygray}97.64 & \cellcolor{myyellow}98.64 & \cellcolor{myyellow}88.73 & \cellcolor{myyellow}99.12 \\
    \hline
    F2F $\rightarrow$ DF,NT,FS & \cellcolor{myyellow}99.49 & \cellcolor{mygray}97.38 & \cellcolor{myyellow}93.02 & \cellcolor{myyellow}98.15 \\
    \hline
    NT $\rightarrow$ DF,F2F,FS & \cellcolor{myyellow}98.90 & \cellcolor{myyellow}97.20 & \cellcolor{mygray}94.52 & \cellcolor{myyellow}95.35 \\
    \hline
    FS $\rightarrow$ DF,F2F,NT & \cellcolor{myyellow}99.26 & \cellcolor{myyellow}99.18 & \cellcolor{myyellow}86.91 & \cellcolor{mygray}99.00 \\
\bottomrule
\end{tabular}
\label{tab:o2o} 
\vspace*{-2mm}
\end{table}

\begin{table*}[t]
\vspace*{-1mm}
\centering
\small
\caption{Comparison of our method with state-of-the-arts under O2O cross-manipulation scenario at HQ level (top) and LQ level (bottom) in FF++. The bold number denotes the best performance and the underlined one denotes the second best.}
\vspace*{-2mm}
\begin{tabular}{l|c|c|c|c|c|c|c|c|c|c|c|c}
\toprule
    \multirow{2}{*}{Method} & \multicolumn{3}{c|}{DF} & \multicolumn{3}{c|}{F2F} & \multicolumn{3}{c|}{NT} &  \multicolumn{3}{c}{FS} \\
		\cline{2-13}
    & F2F & NT & FS & DF & NT & FS & F2F & DF & FS & F2F & NT & DF \\
    \hline \hline
    (HQ) Face X-ray \cite{Li-etal2020learning} & 63.30 & 69.80 & 60.00 & 63.00 &  94.50 & 93.80 & 91.70 & 70.50 & 91.00 &  96.10 & 95.70 & 45.80 \\
    \hline
    (HQ) SBI \cite{shiohara2022detecting} & 67.08 & 62.14 & 72.82 & 91.63 & 62.14 & 72.82 & 67.08 & 91.63 & 72.82 & 67.08 & 62.14 & 91.63 \\
    \hline
    (HQ) LTW \cite{sun2021domain} & 80.02 & 77.30 & 64.00 & 92.70 & 77.30 & 64.00 & 80.02 & 92.70 & 64.00 & 80.02 & 77.30 & 92.70 \\
    \hline
    (HQ) Xception \cite{rossler2019faceforensics} & 73.60 & 73.60 & 49.00 & 80.30 & 69.60 & 76.20 & 81.30 & 80.00  & 73.10 & 88.80 & 71.30 & 66.40 \\
    \hline
    (HQ) SRM \cite{luo2021generalizing} & 76.40 & 81.40 & 49.70 & 83.70 & \textbf{98.40} & \underline{98.70} & \textbf{99.50} & 89.40 & \underline{99.30} & \textbf{99.30} & \bf 98.00 & 68.50 \\
    \hline
    (HQ) SOLA-\textit{sup} \cite{fei2022learning} & \underline{97.29} & \textbf{98.48} & 69.72 & \underline{99.73} & \underline{96.02} & 93.50 & \underline{97.69} & \textbf{99.64} & \textbf{99.76} & 98.13 & 92.07 & \textbf{99.11} \\
    \hline
    (HQ) FTCN \cite{zheng2021exploring} & - & - & - & 98.00 & 96.00 & 95.90 & - & - & - & - & - & - \\
    \hline
    (HQ) FOT \cite{cozzolino2018forensictransfer} & - & - & - & - & - & 72.57 & - & - &- & - & - & - \\
    \hline
    (HQ) DDT \cite{aneja2020generalized} & - &  64.10 & - & - & - & - & - & 78.82 &- & - & - & - \\
    \hline
    (HQ) FEAT \cite{chen2021featuretransfer} & - & - & \underline{88.62} & - & - & - & - & - &- & - & - & - \\
    \hline
    (HQ) TALL-Swin \cite{xu2023tall} & 52.96 & 52.11& 77.36 & 62.55 & 60.49 & 55.54 & 64.92 & 62.09 & 52.20  & 50.55 & 51.96 & 85.08 \\
    \hline
    (HQ) Baseline & 69.43 & 69.55 & 55.18 & 81.71 & 55.84 & 69.74 & 55.19 & 79.74 & 44.98 & 71.06 & 43.63 & 84.34 \\
    \rowcolor{mygray} (HQ) \bf Baseline+BA (Ours) & \textbf{99.08} & \underline{95.03} & \textbf{99.37} & \textbf{99.77} & 95.85 &  \textbf{99.89} & 97.44 & \underline{99.28} & 97.81 & \underline{99.14} & \underline{96.27} & \underline{98.90} \\
    \hline \hline
    (LQ) SBI \cite{shiohara2022detecting} & 70.55 & 65.95 & 68.25 & \underline{85.60} & 65.95 & \underline{68.25} & 70.55 & \underline{85.60} & \underline{68.25} & 70.55 & \underline{65.95} & \underline{85.60} \\
    \hline
    (LQ) FDFL \cite{li2021frequencyaware}  &  58.90 &  63.61 &  66.87 &  67.55 &  55.35 &  66.66 & 74.21 & 79.09 &  74.21 &  54.64 & 49.72 & 75.90 \\
    \hline
    (LQ) LTW \cite{sun2021domain} & \underline{72.40} & 60.80 & 68.10 & 75.60 & 60.80 & 68.10 & 72.40 & 75.60 & 68.10 & \underline{72.40} & 60.80 & 75.60 \\
    \hline
    (LQ) MATT \cite{zhao2021multiattentional} & 66.41 & 66.01 & 67.33 & 73.04 & 71.88 & 65.10 & 80.61 & 74.56 & 60.90 & 61.65 & 54.79 & 82.33 \\
    \hline
    (LQ) RECCE \cite{cao2022endtoend} & 70.66 & \underline{67.34} & \underline{74.29} & 75.99 & \underline{72.32} & 64.53 & \underline{80.89} & 78.83 & 63.07 & 64.44 & 56.70 & 82.39 \\
    \hline
    (LQ) TALL-Swin \cite{xu2023tall} & 55.50 & 52.93 & 74.39 & 63.64 & 57.82 & 53.15 & 68.65 & 63.19 & 51.42 & 50.44 & 53.11 & 83.91 \\
    \hline
   (LQ) Baseline & 61.18 & 59.49 & 71.17 & 65.88 & 51.43 & 59.07 & 56.36 & 65.33 & 52.49 & 56.25 & 44.38 & 70.45 \\
    \rowcolor{mygray} (LQ) \bf Baseline+BA (Ours)  & \textbf{92.13} & \textbf{86.89} & \textbf{98.09} & \textbf{93.52} & \textbf{79.64} & \textbf{94.22}  &  \textbf{86.96} &  \textbf{96.38} & \textbf{93.43} & \textbf{88.16} & \textbf{80.85} & \textbf{99.02} \\
\bottomrule
\end{tabular}
\label{tab:o2ocompare} 
\vspace*{-4mm}
\end{table*}

\begin{figure*}[!bp]
\vspace*{-7mm}
\centering
    \subfloat[{\scriptsize Without ours FS $\rightarrow$ F2F (F2F)}]{
    \includegraphics[width=0.24\linewidth]{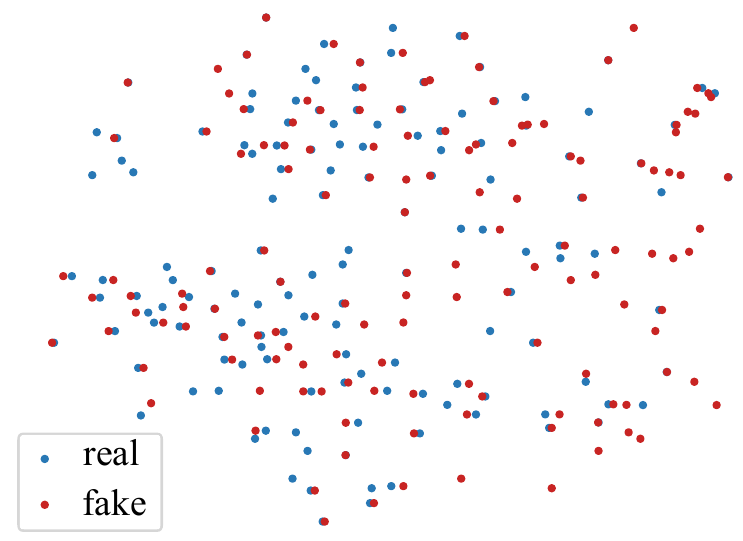}
    }
    \subfloat[{\scriptsize With ours FS $\rightarrow$ F2F (F2F)}]{
    \includegraphics[width=0.24\linewidth]{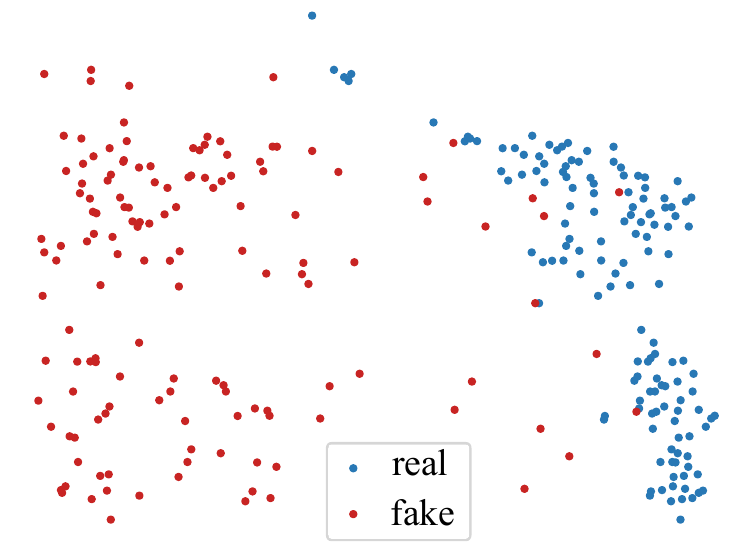}
    }
    \subfloat[{\scriptsize Without ours FS $\rightarrow$ NT (NT)}]{
    \includegraphics[width=0.24\linewidth]{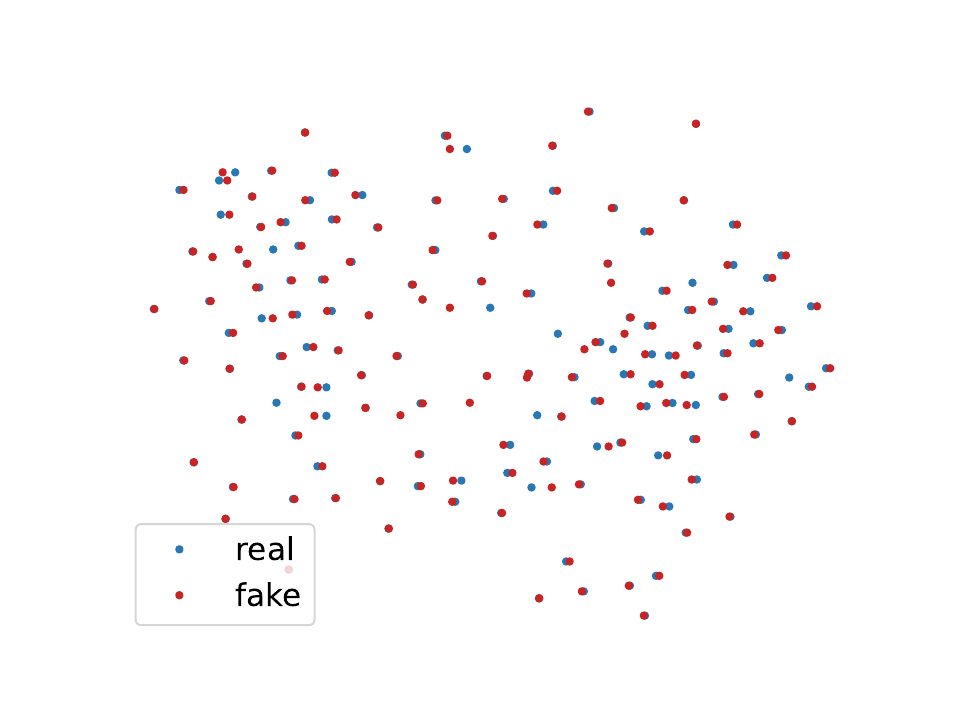}
    }
    \subfloat[{\scriptsize With ours FS $\rightarrow$ NT (NT)}]{
    \includegraphics[width=0.24\linewidth]{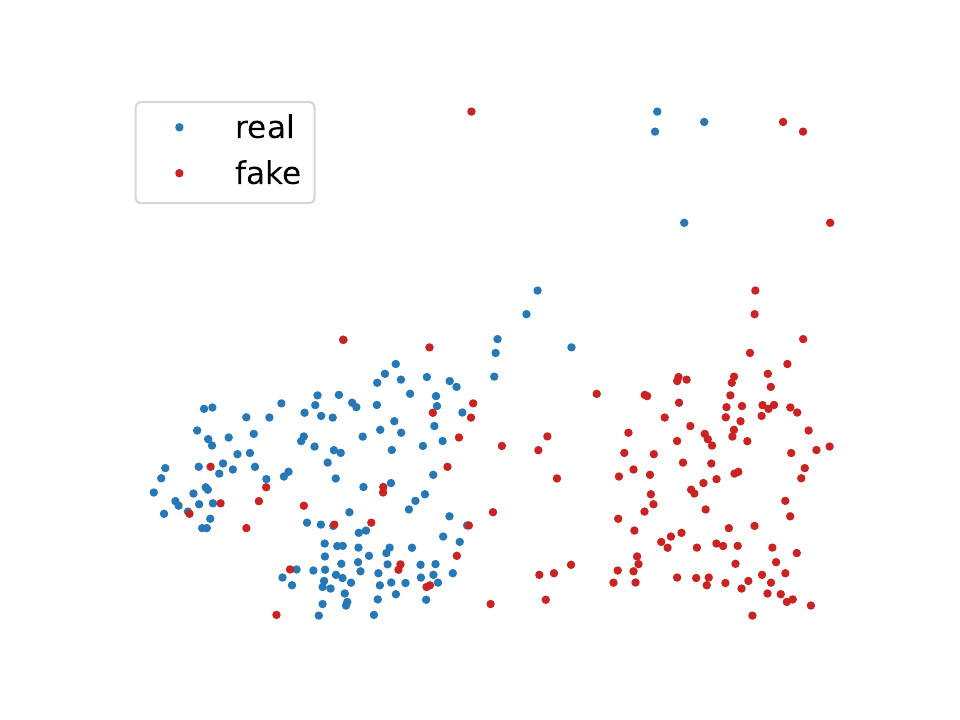}
    }
    
    \subfloat[{\scriptsize Without ours FS $\rightarrow$ F2F (FS)}]{
    \includegraphics[width=0.24\linewidth]{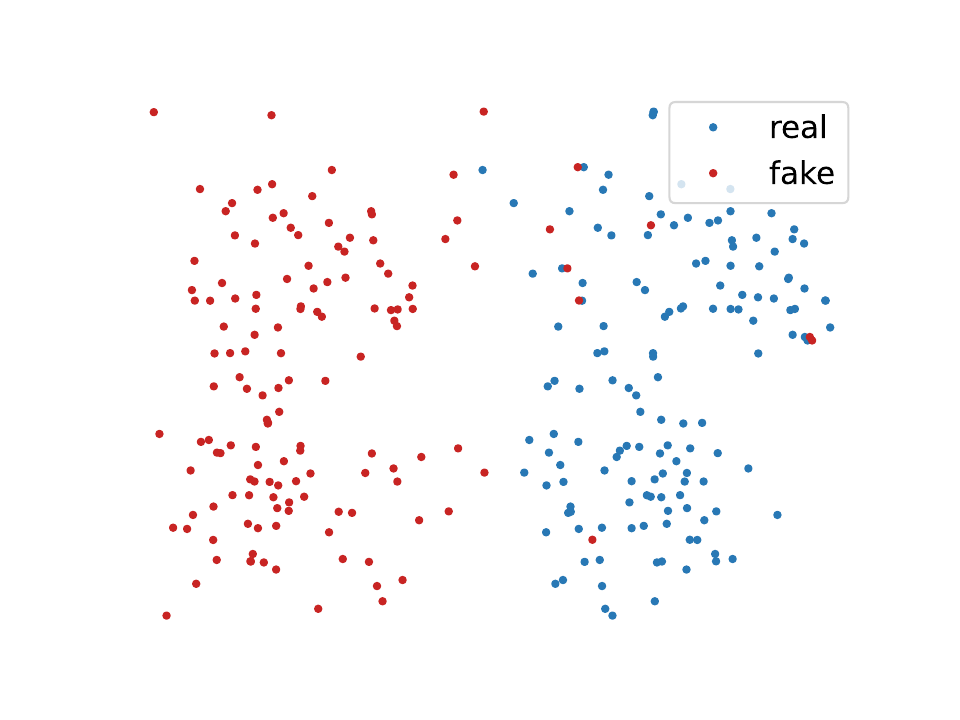}
    }
    \subfloat[{\scriptsize With ours FS $\rightarrow$ F2F (FS)}]{
    \includegraphics[width=0.24\linewidth]{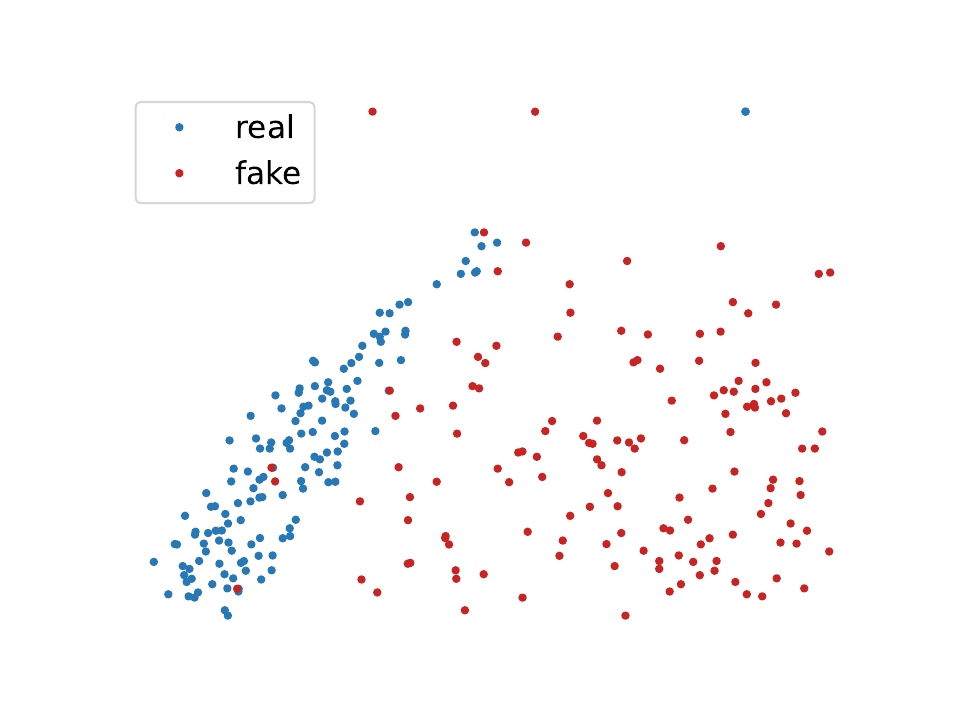}
    }
    \subfloat[{\scriptsize Without ours FS $\rightarrow$ NT (FS)}]{
    \includegraphics[width=0.24\linewidth]{./imgs/tsne_vitso_fs2f2f_fs.pdf}
    }
    \subfloat[{\scriptsize With ours FS $\rightarrow$ NT (FS)}]{
    \includegraphics[width=0.24\linewidth]{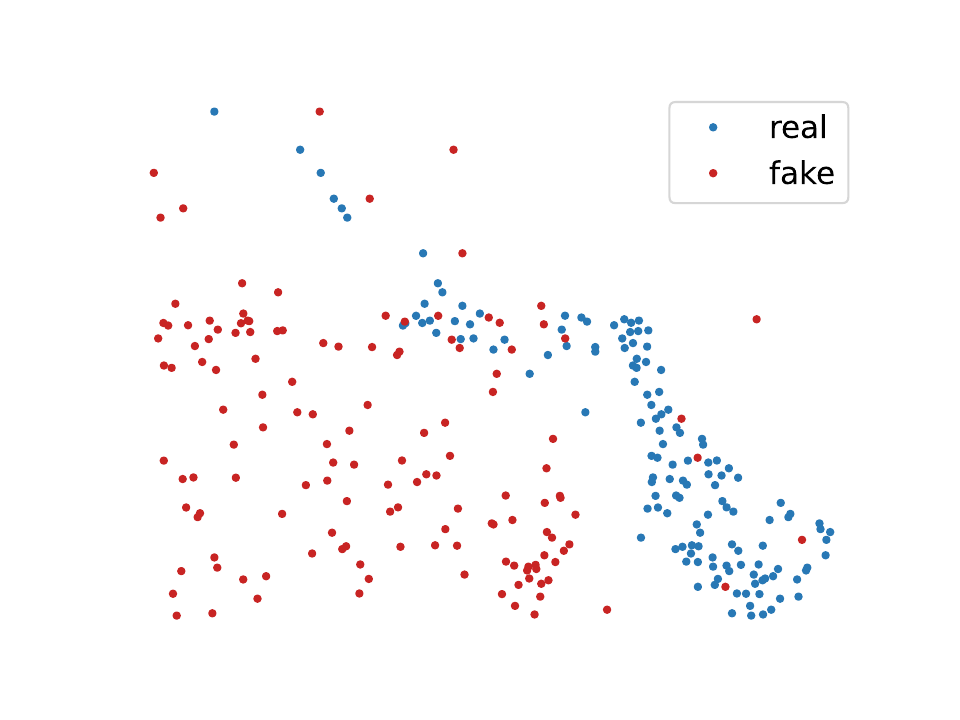}
    }
    
\caption{T-SNE \cite{van2008visualizing} visualization on FF++ (HQ).}
\label{fig:tsneall} 
\vspace{-2mm}
\end{figure*}

\subsubsection{Implementation Details}

Following the design in Vit-Base \cite{dosovitskiy2021an}, the feature extractor contains 12 (i.e., $l=12$) transformer blocks, while we use 4 (i.e., $m=12$) transformer blocks in frequency branch. The input images are resized to $(224, 224)$. The weights of loss terms are set to $\alpha_1=\alpha_2=\alpha_3=\alpha_4=1$. The training batch size is set to $32$ in forward adaptation and $24$ in backward adaptation, as the teacher model occupy extra GPU memory. In the forward adaptation stage, we employ SGD as the optimizer, where the learning rate, momentum, and weight decay are set to $0.001$, $0.9$, and $0.0005$, respectively. We stop the training of forward adaptation after $20$ epochs. For backward adaptation, we set the learning rate to $0.0001$ and the number of training epochs to $10$. The temperature is set to $\tau = 0.5$. All experiments are conducted using Pytorch \cite{pytoch_ARXIV} on two Nvidia GTX 2080Ti GPUs.

\subsection{Results of Cross-manipulation-methods}\label{S4.2}

We investigate two settings in this scenario: one-to-one adaptation (O2O) and one-to-many adaptation (O2M). O2O uses one method as source and another one as target, while O2M uses one method as source and many other methods as target, which is more practical, as the collected target videos may have many new forgeries. All the experiments in this part are conducted on FF++ dataset and evaluated using the area under curve (AUC) metric.

\subsubsection{Performance of O2O and O2M}

In O2O, we select one of the four manipulation methods (DF, F2F, NT, FS) as the source domain and select a different method as the target domain. The top part of Table~\ref{tab:o2o} shows the performance of our method under this O2O setting on the HQ set. Gray and yellow coloured values denote the performance on the source domain and target domain, respectively. ``Baseline'' denotes training the proposed architecture on source domain and directly testing on other domains without bi-directional adaptation. The results reveal that our method performs favourably on both source and target domains, achieving the AUC metric $95\%+$ on most of the source and target adaptation pairs.
This demonstrates the effectiveness of our method. We also observe that the performance with NT $\rightarrow$ FS is limited on the source. This is probably due to the large gap between NT and FS, and thus part of the knowledge in the source domain is lost.  

In O2M, we select one manipulation method as source domain and use the other three methods as target domain. The performance achieved by our method are shown in the bottom part of Table~\ref{tab:o2o}. We can observe that our method performs well on all the manipulation methods, which demonstrates that our method can learn the common forgery features even if the target domain is mixed with different manipulations. 

Compared to the O2O scenario, O2M is more practical, as the daily emerged videos likely contain various manipulation methods. However, the foundation of O2M is O2O, as it still attempts to find the common knowledge among these various manipulation methods. Thus the O2O setting is also important, as it serves as the basis for the improvement of O2M. 

\subsubsection{Comparison with State-of-the-Arts}

We compare our method with several state-of-the-art methods, including the augmentation-based methods (Face X-ray \cite{Li-etal2020learning}, SBI \cite{shiohara2022detecting}), frequency-based methods (SRM \cite{luo2021generalizing}, FDFL \cite{li2021frequencyaware}), transfer-based methods (FOT \cite{cozzolino2018forensictransfer}, DDT \cite{aneja2020generalized}, FEAT \cite{chen2021featuretransfer}), and other methods (Xception \cite{rossler2019faceforensics}, MATT \cite{zhao2021multiattentional}, LTW \cite{sun2021domain}, RECCE \cite{cao2022endtoend}, SOLA-\textit{sup} \cite{fei2022learning}, FTCN \cite{zheng2021exploring}, TALL-Swin \cite{xu2023tall}). Note that TALL-Swin has no reports corresponding to this scenario. Thus we reproduce its results following its default settings. This method is trained using all videos in FF++. 

{\em We would like to clarify that the experiment configurations between our method and these other methods are different, as these methods tackle this task under different scenarios. Specifically, these data-level methods first obtain and empirically analyze the new forgery videos, and then summarize a common forgery knowledge as a prior, e.g., the blending boundary and frequency clues. By contrast, the prior knowledge for us is the collected videos without labels. Even though the experiment configurations are not the same, the results can also reflect the effectiveness of our method in exposing new forgeries.}

Table~\ref{tab:o2ocompare} compares the performance of our method with those of the existing methods under the O2O scenario on both HQ and LQ levels, where the results of our method are marked by gray colour. For the existing methods, we use the score of each method reported in its original paper. As SBI \cite{shiohara2022detecting} was only evaluated on raw quality images, we retrain it using the code provided in \cite{shiohara2022detecting} on HQ and LQ for a fair comparison. 

For the HQ level, it can be seen from the top part of Table~\ref{tab:o2ocompare} that our method outperforms the augmentation-based methods, Face X-ray and SBI, by a large margin. Since these two methods require prior knowledge of manipulation, they can hardly handle the forgery that has notable differences from the prior knowledge. The frequency-based method SRM performs well on several cases, notably, FS~$\rightarrow$~F2F and NT~$\rightarrow$~F2F. But its performances are highly degraded in many more cases compared to our method. For example, in DF~$\rightarrow$~FS, our method is nearly 100\% better than SRM, and in FS~$\rightarrow$~DF, our method is 44\% better than SRM, while in DF~$\rightarrow$~F2F, our method is 30\% better than SRM, and in F2F~$\rightarrow$~DF, our method is 20\% better than SRM. Similar observations can be drawn by comparing our method with SOLA-\textit{sup}. Based on the available data from \cite{zheng2021exploring}, the performance of our method are slightly better than those of FTCN. When compared to TALL-Swin, we observe our method achieves better generalization performances than TALL-Swin. It can also be seen that our method significantly outperforms LTW, Xception, FOT, DDT and FEAT. Moreover, in the more challenging LQ level, our method significantly outperforms all the five benchmarks, as can be seen from the bottom part of Table~\ref{tab:o2ocompare}. 

\begin{figure*}[!t]
\vspace*{-2mm}
\centering
\includegraphics[width=0.8\linewidth]{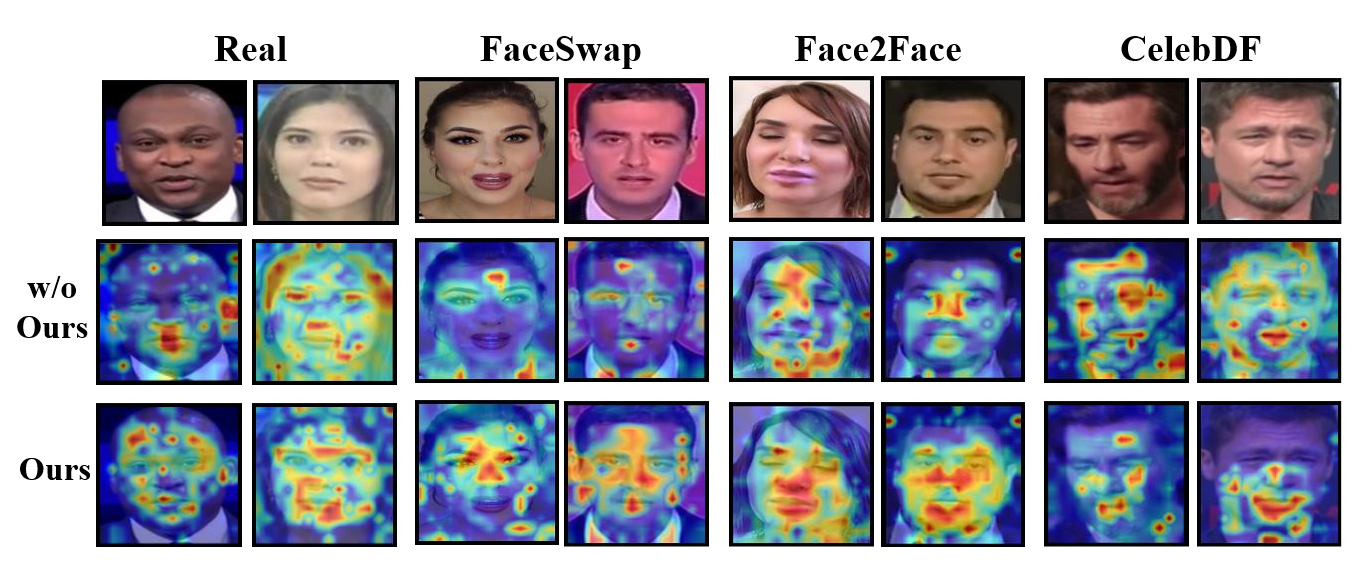}
\vspace{-3mm}
\caption{Grad-CAM \cite{selvaraju2017grad} visualization without and with adaptation using our method.}
\label{fig:gradcam} 
\vspace*{-2mm}
\end{figure*}

\begin{table*}[tp!]
\vspace*{-1mm}
\centering
\small
\caption{Comparison of our method with state-of-the-arts domain adaptation methods. The bold number denotes the best performance.}
\vspace*{-2mm}
\begin{tabular}{l|c|c|c|c|c|c|c|c|c|c|c|c}
\toprule
    \multirow{2}{*}{Method} & \multicolumn{3}{c|}{DF} & \multicolumn{3}{c|}{F2F} & \multicolumn{3}{c|}{NT} &  \multicolumn{3}{c}{FS} \\
		\cline{2-13}
    & F2F & NT & FS & DF & NT & FS & F2F & DF & FS & F2F & NT & DF \\
    \hline \hline
    (HQ) SSRT \cite{sun2022safe_ssrt} & 51.43 & 50.36 & 61.79 & 86.79 & 49.64 & 50.36 & 54.64 & 90.71 & 49.29 & 58.21 & 50.00 & 86.43  \\
    \hline
    (HQ) MDD \cite{zhang2019bridging_mdd}  &  71.79& 73.57 & 61.79 & 80.36 & 70.00 & 80.71 & 77.50 & 88.93 & 75.00 & 75.71 & 50.37 & 73.93 \\
    \hline
    (HQ) DANN \cite{DANN_JMLR_ODA_5}  &  95.71 & 90.71 & 94.29 & 94.99 & 88.57 & 96.43 & 90.36 & 92.86 & 89.64 & 91.43 & 87.14 & 95.00 \\
    \hline
    \rowcolor{mygray} (HQ) \bf Ours & \textbf{99.08} & \textbf{95.03} & \textbf{99.37} & \textbf{99.77} & \textbf{95.85} &  \textbf{99.89} & \textbf{97.44} & \textbf{99.28} & \textbf{97.81} & \textbf{99.14} & \textbf{96.27} & \textbf{98.90} \\
    \hline \hline
    \hline
    (LQ) SSRT \cite{sun2022safe_ssrt}  &  50.35 &  50.40 &  56.79 &  73.21 &  54.28 &  56.43 & 66.79 & 68.93 &  59.29 &  50.36 & 50.14 & 92.14 \\
    \hline
    (LQ) MDD \cite{zhang2019bridging_mdd}  & 62.50 & 57.14 & 62.14 & 68.93 & 56.79 & 62.50 & 70.71 & 77.49 & 65.71 & 59.29 & 49.99 & 77.50\\
    \hline
    (LQ) DANN \cite{DANN_JMLR_ODA_5}  &  78.21 & 71.43 & 83.93 & 82.86 & 66.78 & 81.79 & 77.50 & 81.07 &  75.71 &  80.36 & 68.21 & 90.36\\
    \hline
    \rowcolor{mygray} (LQ) \bf Ours  & \textbf{92.13} & \textbf{86.89} & \textbf{98.09} & \textbf{93.52} & \textbf{79.64} & \textbf{94.22}  &  \textbf{86.96} &  \textbf{96.38} & \textbf{93.43} & \textbf{88.16} & \textbf{80.85} & \textbf{99.02} \\
\bottomrule
\end{tabular}
\label{tab:damethod} 
\vspace*{-4mm}
\end{table*}

Fig.~\ref{fig:tsneall} shows the t-SNE \cite{van2008visualizing} visualizations of feature distribution for two O2O cross-manipulations (FS $\rightarrow$ F2F, FS $\rightarrow$ NT) without and with adaptation using our method on both source and target domain. For the target domain, it can be seen that the features of real and fake faces separate well using our method, in contrast to the mixed features without our method (see (a,b) and (c,d)). For the source domain, we can observe that their feature distributions are still discriminative (see (e,f) and (g,h)). Fig.~\ref{fig:gradcam} shows several examples of Grad-CAM \cite{selvaraju2017grad} (the left six columns), which reveals that our method catches more discriminative features on face regions for both real and fake faces. 

\subsubsection{Comparison with Existing UDA Methods}  

As our approach draws inspiration from UDA to enhance the generalization ability when facing unseen DeepFake techniques, we also perform a further comparison with the recently developed domain adaptation methods, SSRT \cite{sun2022safe_ssrt}, MDD \cite{zhang2019bridging_mdd} and DANN \cite{DANN_JMLR_ODA_5}. Specifically, we retrain their models on FF++ HQ and LQ scenarios by using their published codes. The experimental results are summarized in Table~\ref{tab:damethod}, which demonstrate that our method outperforms these three domain adaptation methods in recognizing faces with unseen forgeries. SSRT \cite{sun2022safe_ssrt} is a counterpart method that also employs a powerful vision transformer as the backbone network. It can be seen that SSRT occasionally fails to transfer knowledge in some scenarios, e.g., DF$\rightarrow$NT and FS$\rightarrow$F2F, while our bi-directional network achieves more stable and superior performance than SSRT almost in all adaptation tasks. DANN and MDD respectively represent an adversarial-based method and a discrepancy-based method. We observe that DANN achieves more generalized performance than MDD but our approach surpasses both DANN and MDD by a large margin. The reason mainly lies in that the proposed bi-directional adaptation strategy enables the model to fully grab the common forgery features across domains. Overall, the results suggest that our approach effectively encourages the detector to learn common forgery features, leading to improved generalization ability.

\begin{table}[!bh]
\vspace*{-6mm}
\centering
\small
\caption{Performance comparison of our method with state-of-the-arts under cross-datasets scenario from FF++ to Celeb-DF. Training set FF++ indicates FF++ with four manipulation methods (DF,F2F,NT,FS) as source, and FF++(DF) means FF++ with only one manipulation method DF as the training set.}
\vspace*{-1mm}
\resizebox{\linewidth}{!}{
\begin{tabular}{l|c|c|c}
    \toprule
    \multirow{2}{*}{Method} & \multirow{2}{*}{Training Set}  & \multirow{2}{*}{FF++} & Testing Set \\
    \cline{4-4}
    & & & Celeb-DF \\
    \hline \hline
    (HQ) LTW \cite{sun2021domain}                  & FF++ & - & 64.10 \\
    \hline
    (LQ) MATT \cite{zhao2021multiattentional}      & FF++ & 99.80 & 67.02 \\
    \hline
    (HQ) DSP-FWA \cite{li2019exposing}             & FF++ & - & 69.30 \\
    \hline
    (HQ) MTD-Net  \cite{TIFSnew1}                  & FF++ & 99.38 & 70.12 \\
    \hline
    (HQ) Xception \cite{rossler2019faceforensics}  & FF++ & 95.73 & 73.04 \\
    \hline
    (HQ) CFFs \cite{TIFSnew2}                      & FF++ & 97.03 & 74.20 \\
    \hline
    (HQ) DAM \cite{zhou2021faceinwild}             & FF++ & - & 75.30 \\
    \hline
    (HQ) SOLA-\textit{sup} \cite{fei2022learning}  & FF++(DF) & - & 76.02 \\
    \hline
    (HQ) SPSL \cite{liu2021spatial}           & FF++  & 95.32 & 76.88 \\
    \hline
    (HQ) LiSiam  \cite{TIFSnew3}                   & FF++ & 99.13 & 78.21 \\
    \hline
    (HQ) SLADD \cite{chen2022selfsupervised}       & FF++ & 98.40 & 79.70 \\
    \hline
    (HQ) LipForensics \cite{haliassos2021lips}     & FF++ & 99.70 & 82.40 \\
    \hline
    (HQ) HFI-Net \cite{TIFSnew4}                   & FF++ & 98.86 & 83.28 \\
    \hline
    (HQ) TALL-ViT \cite{xu2023tall}     & FF++ & - & 86.58 \\
    (HQ) TALL-Swin \cite{xu2023tall}     & FF++ & - & \textbf{90.79} \\
    \hline
    (HQ) FTCN \cite{zheng2021exploring}            & FF++ & - & 86.90 \\
    \hline
    (HQ) AltFreezing \cite{wang2023altfreezing}    & FF++ & 99.70 & 89.50 \\
    \hline
    (HQ) F$^2$Trans \cite{TIFSnew5}                & FF++ & 99.74 & 89.87 \\
    \hline
    (RAW) PCL+I2G \cite{zhao2021learning}          & FF++ & 99.79 & 90.03 \\
    \hline
    \rowcolor{mygray} (HQ) Baseline            & FF++ & 99.13 & 81.90 \\
    \rowcolor{mygray} (HQ) \bf Baseline+BA (Ours)  & FF++ & 98.64 & 90.22 \\
    \hline \hline
    (LQ) F$^3$-Net \cite{qian2020thinking}         & FF++ & 93.30 & 61.51 \\
    \hline
    (LQ) RECCE \cite{cao2022endtoend}              & FF++ & 95.02 & 68.71 \\
    \hline
    (LQ) Two-branch \cite{twobranch}               & FF++ & 93.18 & 73.41 \\
    \hline
    \rowcolor{mygray} (LQ) Baseline            & FF++ & 94.25 & 75.68\\
    \rowcolor{mygray} (LQ) \bf Baseline+BA (Ours)  & FF++ & 91.24 & \textbf{81.39} \\
\bottomrule
\end{tabular}}
\label{tab:one2celeb} 
\vspace*{-1mm}
\end{table}

\subsection{Results of Cross-manipulation-datasets}\label{S4.3}

In this experiment, we use FF++ as source domain and Celeb-DF as target domain. Table~\ref{tab:one2celeb} compares our method with the state-of-the-art methods, F$^3$-Net \cite{qian2020thinking}, MATT \cite{zhao2021multiattentional}, RECCE \cite{cao2022endtoend}, Two-branch \cite{twobranch}, LTW \cite{sun2021domain}, DSP-FWA \cite{li2019exposing}, MTD-Net \cite{TIFSnew1}, Xception \cite{rossler2019faceforensics}, CFFs \cite{TIFSnew2}, DAM \cite{zhou2021faceinwild}, SOLA-\textit{sup} \cite{fei2022learning}, SPSL \cite{liu2021spatial}, LiSiam \cite{TIFSnew3}, SLADD \cite{chen2022selfsupervised}, LipForensics \cite{haliassos2021lips}, HFI-Net \cite{TIFSnew4}, FTCN \cite{zheng2021exploring}, AltFreezing \cite{wang2023altfreezing}, TALL \cite{xu2023tall}, F$^2$Trans \cite{TIFSnew5}, and PCL+I2G \cite{zhao2021learning}. It can be seen from Table~\ref{tab:one2celeb} that our method achieves $90.22\%$ at HQ and $81.39\%$ at LQ, respectively, outperforming most other methods by a large margin. Note that PCL+I2G \cite{zhao2021learning} synthesizes forged faces with raw quality data since the uncompressed data can provide more distinct features than the compression ones. 
Our method outperforms TALL-ViT \cite{xu2023tall}, which uses spatio-temporal modeling for learning local and global contextual deepfake patterns, when employing a vision transformer as the backbone network. Additionally, it achieves comparable performance with TALL-Swin.
This demonstrates that our method is capable of transferring the forgery knowledge from one dataset to another. The last two columns of Fig.~\ref{fig:gradcam} depict the examples of Grad-CAM from FF++ to Celeb-DF, which shows that our method concentrates more on forgery regions than the case without using our method.

For a comprehensive study, we then conduct a more challenging scenario, which uses the FF++ as the source domain and the DFDCP \cite{dfdcp_dataset} or FFIW \cite{zhou2021faceinwild} as the target domain. The results are presented in Table~\ref{tab:ff2dfdcp_ffiw}. We observe that our method achieves better performance than others on the DFDCP dataset, mainly because our method can learn common forgery features among different manipulated methods. However, SBI surpasses our method on the FFIW dataset. This is because this dataset contains many side faces, while the faces in FF++ dataset are usually frontal. This discrepancy largely increases the domain gap, which may disturb our model in learning common forgery features. Since SBI is designed to create a variety of faces for data augmentation, which likely includes such distorted faces in training, resulting in better performance than us.

\begin{table}[!t]
\vspace*{-2mm}
\centering
\small
\caption{Performance comparison of our method with state-of-the-arts under cross-datasets scenario from FF++ to DFDCP and FFIW.}
\vspace*{-1mm}
\resizebox{0.9\linewidth}{!}{
\begin{tabular}{l|c|c|c}
\toprule
    Method & Training Set & DFDCP \cite{dfdcp_dataset} & FFIW \cite{zhou2021faceinwild}  \\
    \hline \hline
    PCL+I2G \cite{zhao2021learning}   & FF++(RAW) & 74.37 & -\\
    \hline
    SRM \cite{luo2021generalizing} & FF++(HQ) & 79.70 & - \\
    \hline
    SBI \cite{shiohara2022detecting} & FF++(RAW) & 86.15 & \textbf{84.83}  \\
    \hline
    SBI \cite{shiohara2022detecting} & FF++(HQ) & 85.51 & 83.22  \\
    \hline
    Baseline & FF++(HQ) & 80.37 & 63.85  \\
    \rowcolor{mygray} \bf Baseline+BA (Ours) & FF++(HQ) & \textbf{89.57} & 76.76 \\
\bottomrule
\end{tabular}}
\label{tab:ff2dfdcp_ffiw}
\vspace*{-2mm}
\end{table}

\begin{table}[!h]
\vspace*{-2mm}
\centering
\small
\caption{Performance comparison of our method with state-of-the-arts under cross-faceswap\&GAN scenario from FF++ to StyleGAN.}
\vspace*{-1mm}
\resizebox{\linewidth}{!}{
\begin{tabular}{l|c|c|c|c|c}
\toprule
    Method & DF & F2F & NT & FS & Avg. \\
    \hline \hline
    (HQ) MATT \cite{zhao2021multiattentional} & 43.99 & 43.99 & 43.99 & 43.99 & 43.99 \\
    \hline
    (HQ) RECCE \cite{cao2022endtoend} & 44.80 & 44.80 & 44.80 & 44.80 & 44.80 \\
    \hline
    (HQ) SBI \cite{shiohara2022detecting} & 48.60 & 48.60 & 48.60 & 48.60 & 48.60 \\
    \hline
    \rowcolor{mygray} (HQ) \bf Baseline+BA (Ours) & \textbf{68.55} & \textbf{58.06} & \textbf{55.10} & \textbf{59.08} & \textbf{60.20} \\
    \hline \hline
    (LQ) MATT \cite{zhao2021multiattentional} & 32.51 & 32.51 & 32.51 & 32.51 & 32.51 \\
    \hline
    (LQ) RECCE \cite{cao2022endtoend} & 58.95 & 58.95 & 58.95 & 58.95 & 58.95 \\
    \hline
    (LQ) SBI \cite{shiohara2022detecting} & 64.11 & \textbf{64.11} & 64.11 & 64.11 & 64.11 \\
    \hline
    \rowcolor{mygray} (LQ) \bf Baseline+BA (Ours) & \textbf{71.07} &  52.51 & \textbf{88.62} & \textbf{70.28} & \textbf{70.62} \\
\bottomrule
\end{tabular}}
\label{tab:one2stylegan} 
\vspace*{-2mm}
\end{table}

\subsection{Results of Cross-manipulation-types}\label{S4.4}

In this part, we investigate the feasibility of our method to adapt faceswap faces to GAN-synthesized faces. Faceswap replaces the central region of face and retains other regions unchanged, while GAN synthesizes the whole face image. We use each of faceswap methods in FF++ \cite{rossler2019faceforensics} as source and adapt it to the StyleGAN dataset \cite{stylegan}. Table \ref{tab:one2stylegan} shows the performance of our method in comparison to the three benchmark methods, MATT \cite{zhao2021multiattentional}, RECCE \cite{cao2022endtoend} and SBI \cite{shiohara2022detecting}, under this setting. For fair comparison, we use the codes provided in \cite{zhao2021multiattentional, cao2022endtoend, shiohara2022detecting} for SBI, MATT and RECCE under their defaulting settings. Observe that these three methods perform poorly, and in most cases their performance are below 50\%. This is because these methods are designed to detect faceswap DeepFakes, e.g., finding the blending artifacts, and they cannot handle StyleGAN synthesized faces. In contrast, our method outperforms these methods by a large margin, on average 37\%, 34\% and 24\% better at HQ level, and 117\%, 20\% and 10\% better at LQ level, than MATT, RECCE and SBI, respectively. This is because our method focuses on learning the common forgery knowledge in the central regions of both faceswap and GAN DeepFakes, instead of simply exposing the faceswap-specific forgery features.

\subsection{Ablation Study}\label{S4.5}

We also conduct comprehensive ablation experiments to fully analyze the proposed method. More specifically, the ablation study 1)~validates the efficacy of proposed approach; 2)~demonstrates the proposed method is data-efficient and can work with various backbone networks; and 3)~shows that our method can also benefit other augmentation-based face forgery detection methods to improve their generalization ability.

\subsubsection{Various Adaptation Settings}

To provide more insights, we further investigate the proposed bi-directional adaptation strategy on the LQ set of FF++. We use the DeepFake detector trained on the source domain without adaptation as the baseline. GRL \cite{ganin2015unsupervised} and MMD \cite{long2015learning} are two classical domain adaptation methods. GRL uses the gradient reversal layer and MMD attempts to reduce the distance of the probability distributions between source and target domains. GRL is exactly the domain adaptation method adopted in FEAT \cite{chen2021featuretransfer}. We denote the forward adaptation method in our method as FA. In the top part of Table~\ref{tab:comp_abla}, we evaluate these three domain adaptation methods. It can be seen that adding GRL, MMD or FA can improve the performance in the target domain, and using our FA attains the highest performance gain. 

\begin{table}[!t]
\centering
\small
\caption{Effect of various adaptation settings.}
\vspace*{-1mm}
\resizebox{\linewidth}{!}{
\begin{tabular}{l|c|c|c|c|c|c}
\toprule
    \multirow{2}{*}{Method} & \multicolumn{2}{c|}{DF $\rightarrow$ F2F} & \multicolumn{2}{c|}{DF $\rightarrow$ FS} & \multicolumn{2}{c}{NT $\rightarrow$ FS} \\
    \cline{2-7}
    & DF & F2F & DF & FS & NT & FS \\
    \hline
    \hline
    Baseline & 99.32 & 57.98 & 99.53 & 76.18 & 87.61 & 53.45 \\
    \hline \hline
    + GRL \cite{ganin2015unsupervised} & 99.29 & 68.32 & 99.55 & 89.46 & 88.22 & 71.11 \\
    \hline  
    + MMD \cite{long2015learning} & 99.22 & 84.12 & 99.52 & 94.76 & 87.63 & 83.88 \\
    \hline
    + FA & 99.30 & 86.49 & 99.62 & 95.61 & 87.90 & 84.53 \\
    \hline \hline
    + SD & 97.85 & 90.59 & 96.02 & 98.57 & 77.53 & 95.89 \\
    \hline
    + Ent. \cite{entropy} & 74.89 & 54.98 & 68.73 & 54.92 & 55.67 & 50.67 \\
    \hline
    Baseline+BA (Ours) & 96.43 & 92.13 & 98.08 & 98.09 & 85.03 & 93.94 \\
\bottomrule
\end{tabular}}
\label{tab:comp_abla} 
\vspace*{-4mm}
\end{table}

\begin{figure}[!b]
\vspace*{-5mm}
\centering
\includegraphics[width=\linewidth]{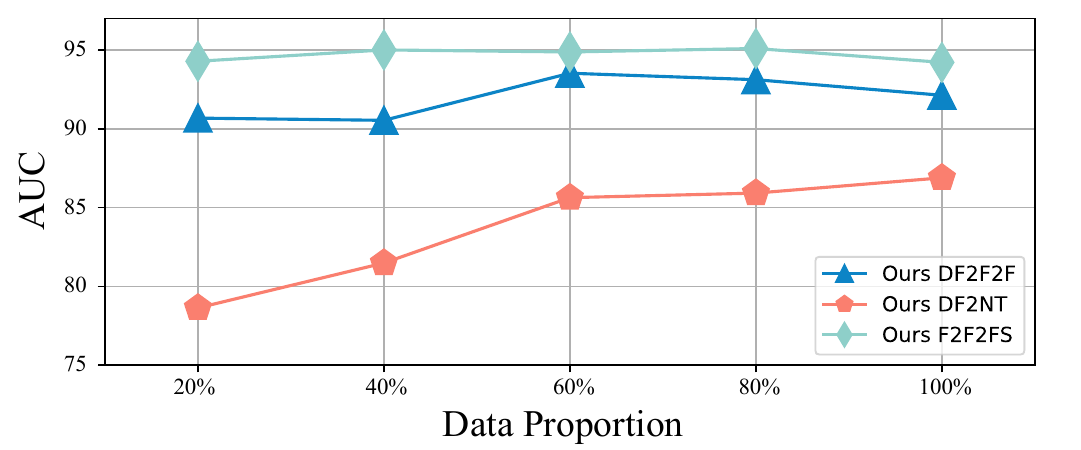}
\vspace*{-7mm}
\caption{Performance of using various amounts of training samples.}
\label{fig:abla_dataratio} 
\vspace*{-1mm}
\end{figure}

Then we take a further step to analyze the effect of backward adaptation in the bottom part of Table~\ref{tab:comp_abla}, where SD means that the self-distillation is used in backward adaptation, while Ent. represents that self-distillation is replaced with entropy minimization of samples, which is inspired by the method of \cite{entropy}. We observe that only adding SD can notably improve the performance on the target domain, but its performance on the source domain is compromised, especially in NT~$\rightarrow$~FS, with $77.53\%$ of +SD v.s. $87.61\%$ of Baseline. This is because self-distillation is effective on the target domain, but overlooks the adaptation learned in forward adaptation. It can also be seen that replacing SD with Ent. is a bad idea, as this strategy degrades the detection performance in both domains. By adding both our forward adaptation and backward adaptation components to the baseline, our method can reach a good balance of the detection performance in both source and target domains. This clearly demonstrates that the proposed bi-directional adaptation is highly effective.

\subsubsection{Various Amounts of Training Samples}

This part investigates the data efficiency of the proposed method. We randomly select a proportion or percentage of the samples in target domain for training to simulate data-constrained scenarios. Fig.~\ref{fig:abla_dataratio} depicts the performance of our method for the adaptation pairs of DF~$\rightarrow$~F2F, DF~$\rightarrow$~NT, and F2F~$\rightarrow$~FS on the LQ set of FF++ using the percentage of the target samples ranging from 20\% to 100\%. As expected, increasing the percentage generally improves achievable performance. More importantly, noting the baseline+FA performance of 86.49 for DF~$\rightarrow$~F2F from Table~\ref{tab:comp_abla}, it can be seen that our method can still considerably enhance the performance to over 90 even if only 20\% of target samples are used. This indicates that our method can be utilized in the more restricted cases where the available target samples are insufficient.

\begin{table}[!h]
\vspace*{-2mm}
\caption{Effect of frequency module.}
\vspace*{-2mm}
\centering
\small
\resizebox{\linewidth}{!}{
\begin{tabular}{l|c|c|c|c|c|c}
\toprule
    \multirow{2}{*}{Method} & \multicolumn{2}{c|}{DF $\rightarrow$ F2F} & \multicolumn{2}{c|}{DF $\rightarrow$ FS} & \multicolumn{2}{c}{NT $\rightarrow$ FS} \\
    \cline{2-7}
    & DF & F2F & DF & FS & NT & FS \\
    \hline
    C-192,D-4 & 93.34 & 90.36 & 96.56 & 97.51 & 80.97 & 92.69 \\
    \hline
    C-384,D-4 & 95.26 & 90.65 & 97.82 & 97.66 & 84.13 & 93.46 \\
    \hline
    C-768,D-2 & 95.17 & 90.97 & 96.97 & 97.35 & 84.21 & 92.39 \\
    \hline
    C-768,D-3 & 95.70  & 91.21 & 97.86 & 96.87 & 84.23 & 92.78 \\
    \hline \hline
    C-768,D-4 & 96.43 & 92.13 & 98.08 & 98.09 & 85.03 & 93.94 \\
\bottomrule
\end{tabular}}
\label{tab:freq_abla} 
\vspace*{-1mm}
\end{table}

\subsubsection{Effect of Frequency Module} 
 
The number of channels (C) and depth (D) of DCT blocks are two important factors for extracting frequency features. The C and D values used in our method are 768 and 4. We now investigate the achievable performance of our method using various C and D values on the LQ set of FF++.  As can be seen from Table~\ref{tab:freq_abla}, given depth $\text{D}=4$, increasing channels from $\text{C}=192$, 384 to 768 improves the achievable performance on both source and target domains. Likewise, with channels fixed to $\text{C}=768$, increasing depth from $\text{D}=2$, 3 to 4 improves the achievable performance on both domains. The results of Table~\ref{tab:freq_abla} also validate that our choice of $\text{C}=768$ and $\text{D}=4$ is appropriate.

\begin{table}[!h]
\vspace*{-2mm}
\centering
\small
\caption{Performance of using various feature extractors.}
\vspace*{-2mm}
\resizebox{\linewidth}{!}{
\begin{tabular}{l|c|c|c|c|c|c}
\toprule
    \multirow{2}{*}{Method} & \multicolumn{2}{c|}{DF $\rightarrow$ F2F} & \multicolumn{2}{c|}{DF $\rightarrow$ FS} & \multicolumn{2}{c}{NT $\rightarrow$ FS} \\
    \cline{2-7}
    & DF & F2F & DF & FS & NT & FS \\
    \hline \hline
    ResNet-50 \cite{resnet_CVPR} & 99.40 & 57.51 & 99.40 & 65.72 & 87.72 & 57.27 \\
    \hline
    ResNet-50 + Ours & 96.89 & 76.17 & 96.49 & 90.61 & 77.99 & 71.21 \\
    \hline \hline
    Xception \cite{chollet2017xception} & 98.71 & 63.04 & 98.71 & 65.11 & 89.81 & 51.93 \\
    \hline
    Xception + Ours & 98.66 & 78.52 & 98.60 & 90.37 & 88.89 & 85.18 \\
    \hline \hline
    EfficientNet-B0 \cite{tan2019efficientnet} & 98.98 & 58.56 & 98.98 &  68.93 & 87.48 & 48.28 \\
    \hline
    EfficientNet-B0 + Ours & 90.26 & 91.24 & 98.32 & 94.66 & 84.45 & 91.57 \\
    \hline \hline
    EfficientNet-B4 \cite{tan2019efficientnet} & 98.82 & 57.91 & 98.82 & 68.63 & 88.77 & 48.56 \\
    \hline
    EfficientNet-B4 + Ours & 93.98 & 88.44 & 99.26 & 90.17 & 89.62 & 90.35 \\
    \hline \hline
    ViT-Small \cite{dosovitskiy2021an} & 98.69 & 59.91 & 98.69 & 75.75 & 86.82 & 52.99 \\
    \hline
    ViT-Small + Ours & 95.63 & 87.46 & 97.27 & 95.58 & 83.83 & 83.17 \\
    \hline \hline
    ViT-Base \cite{dosovitskiy2021an} & 99.32 & 57.98 & 99.53 & 76.18 & 87.61 & 53.45 \\
    \hline
    ViT-Base + Ours & 96.43 & 92.13 & 98.08 & 98.09 & 84.24 & 93.94 \\
\bottomrule
\end{tabular}}
\label{tab:backbone_abla} 
\vspace*{-1mm}
\end{table}

\subsubsection{Various Feature Extractors} 

As mentioned in the introduction section, the proposed framework is independent of the architecture. Table~\ref{tab:backbone_abla} shows the performance of applying our framework on ResNet-50 \cite{resnet_CVPR}, Xception \cite{chollet2017xception}, EfficientNet-B0 \cite{tan2019efficientnet}, EfficientNet-B4 \cite{tan2019efficientnet}, ViT-small \cite{dosovitskiy2021an} and ViT-base \cite{dosovitskiy2021an} on the LQ set of FF++. It can be seen that for each base network, our method significantly improves the performance on the target domain while maintaining a favorable performance on the source domain. This demonstrates that our proposed method is generically applicable.

\subsubsection{Added on State-of-the-art DeepFake Detection} 

This part investigates the effectiveness of our method to improve recently developed advanced detection methods. We use SBI \cite{shiohara2022detecting} as an example. Specifically, we retrain SBI with HQ and LQ faces in FF++, and compare it with our method added on SBI. In this experiment, the source domain consists of fake faces blended by SBI and pristine faces. The target domain is each manipulation method. Thus, four new adaptation sub-tasks named SBI$\rightarrow$ DF, SBI$\rightarrow$ F2F, SBI$\rightarrow$ NT, and SBI$\rightarrow$ FS are formed.
The results shown in Table~\ref{tab:sbi2FF++} reveal that with our method added on, SBI notably improves the performance both at HQ and LQ levels. This indicates that our method can be easily integrated into other detection methods to enhance their generalization ability.

\begin{table}[!h]
\vspace{-2mm}
\centering
\small
\caption{\small Performance of adding our method on SBI.}
\vspace{-2mm}
\resizebox{0.9\linewidth}{!}{
\begin{tabular}{l|c|c}
\toprule
    Adaptation Setting & SBI \cite{shiohara2022detecting} & SBI + Ours \\
    \hline
    \hline
    (HQ) SBI $\rightarrow$ DF  & 85.88 & 99.00 \\
    \hline
    (HQ) SBI $\rightarrow$ F2F  & 80.07 & 97.23 \\
    \hline
    (HQ) SBI $\rightarrow$ NT  & 72.47 & 90.28 \\
    \hline
    (HQ) SBI $\rightarrow$ FS  & 75.62 & 92.92 \\
    \hline
    Avg.  & 78.51 & \bf 94.86 \\
    \hline
    \hline
    (LQ) SBI $\rightarrow$ DF  & 80.70 & 88.93 \\
    \hline
    (LQ) SBI $\rightarrow$ F2F  & 67.63 & 68.20 \\
    \hline
    (LQ) SBI $\rightarrow$ NT  & 65.13 & 64.98 \\
    \hline
    (LQ) SBI $\rightarrow$ FS  & 62.50 & 65.52 \\
    \hline
    Avg.& 68.99 & \bf 71.91 \\
\bottomrule
\end{tabular}}
\label{tab:sbi2FF++} 
\vspace{-1mm}
\end{table}

\subsubsection{Temperature $\tau$} 
We conduct experiments to investigate the influence of temperature in Eq.~\ref{eq4}. As seen in Table~\ref{tab:temp_abla}, we observe that the model training with a large $\tau$ (e.g. $>$ 0.5) achieves stable performance. However, when $\tau$ is set less than 0.5, the result will get unstable.

\begin{table}[!h]
\vspace*{-2mm}
\caption{Effect of temperature $\tau$.}
\vspace*{-2mm}
\centering
\small
\resizebox{0.8\linewidth}{!}{
\begin{tabular}{l|c|c|c|c}
\toprule
    Sub-task  & 0.35 & 0.5 & 0.65 & 0.8 \\
    \hline
    (HQ) DF$\rightarrow$ F2F  & 98.56 & 99.08 & 99.03 & 99.07  \\
    \hline
    (HQ) FS$\rightarrow$ NT & 94.43 & 96.27& 95.60  & 94.25 \\
\bottomrule
\end{tabular}}
\label{tab:temp_abla} 
\vspace*{-1mm}
\end{table}

\subsubsection{Teacher model updating strategy} 
To validate the effectiveness of updating strategy, we conduct addition experiment using exponential moving average (EMA) updating strategy \cite{tarvainen2017meanteacher} on two sub-tasks (e.g. FS$\rightarrow$NT and DF$\rightarrow$F2F). The results are presented in Table~\ref{tab:update_abla}. Our method achieves better performances than exponential moving average on both two sub-tasks. We conjecture this is because the direct assignment makes the knowledge transferred from teacher model to student model faster, which thereby conveys more comprehensive knowledge than EMA strategy.

\begin{table}[!h]
\vspace*{-2mm}
\caption{Effect of teacher model updating strategy.}
\vspace*{-2mm}
\centering
\small
\resizebox{0.8\linewidth}{!}{
\begin{tabular}{l|c|c}
\toprule
    Strategy & (HQ) DF $\rightarrow$ F2F & (HQ) FS $\rightarrow$ NT  \\
    \hline
    EMA \cite{tarvainen2017meanteacher} & 98.49 & 94.19  \\
    \hline
    Baseline+BA (Ours) & 99.08 & 96.27  \\
\bottomrule
\end{tabular}}
\label{tab:update_abla} 
\vspace*{-1mm}
\end{table}

\subsubsection{Results on unseen dataset}

We further conduct an experiment to evaluate the generalization performance on unseen dataset, where we make a comparison on two adaptation settings named FF++(HQ) $\rightarrow$ CelebDF and FF++(HQ) $\rightarrow$ DFDCP. As shown in Table~\ref{tab:ff2dfdcp_celebval}, We observe that our method can also achieve comparable generalization performance on Celeb-DF when the target domain is DFDCP.

\begin{table}[!h]
\vspace*{-2mm}
\centering
\small
\caption{Cross-dataset generalization comparison.}
\vspace*{-1mm}
\begin{tabular}{l|c}
\toprule
    Adaptation Setting  & CelebDF  \\
    \hline \hline
    FF++(HQ) $\rightarrow$ CelebDF & 90.22 \\
    \hline
    FF++(HQ) $\rightarrow$ DFDCP & 85.24 \\
\bottomrule
\end{tabular}
\label{tab:ff2dfdcp_celebval}
\vspace*{-6mm}
\end{table}

\section{Conclusions}\label{S5}

This paper has proposed a new framework to detect new forgeries across different domains, called {\em DomainForensics}. Different from the recent methods, which empirically seeks the common traces on data view, our method aims to transfer the forgery knowledge from known forgeries (fully labeled source domain) to new forgeries (label-free target domain). Since the general domain adaptation methods are not competent to capture the forgery features, we have designed a new bi-directional adaptation strategy that considers both the forward adaptation and backward adaptation. Specifically, the forward adaptation transfers the knowledge from the source to the target domain, and the backward adaptation reverses the adaptation from the target to the source domain. With this backward adaptation, the detector can be further enhanced to learn new forgery features from unlabeled data and avoid forgetting the known knowledge of known forgery. Extensive experiments have been conducted on three datasets with the comparison to several existing state-of-the-art counterparts. The results obtained have demonstrated that our method is effective in exposing new forgeries, and it can be integrated into various architectures to improve their generalization ability. 

\section{Acknowledgement}
This work is supported by National Key Research and Development Program of China under grant number 2022ZD0117201. Yuezun Li is supported by China Postdoctoral Science Foundation under grant number 2021TQ0314 and grant number 2021M703036.

\bibliographystyle{IEEEtran}

\begin{thebibliography}{10}
\providecommand{\url}[1]{#1}
\csname url@samestyle\endcsname
\providecommand{\newblock}{\relax}
\providecommand{\bibinfo}[2]{#2}
\providecommand{\BIBentrySTDinterwordspacing}{\spaceskip=0pt\relax}
\providecommand{\BIBentryALTinterwordstretchfactor}{4}
\providecommand{\BIBentryALTinterwordspacing}{\spaceskip=\fontdimen2\font plus
\BIBentryALTinterwordstretchfactor\fontdimen3\font minus
  \fontdimen4\font\relax}
\providecommand{\BIBforeignlanguage}[2]{{%
\expandafter\ifx\csname l@#1\endcsname\relax
\typeout{** WARNING: IEEEtran.bst: No hyphenation pattern has been}%
\typeout{** loaded for the language `#1'. Using the pattern for}%
\typeout{** the default language instead.}%
\else
\language=\csname l@#1\endcsname
\fi
#2}}
\providecommand{\BIBdecl}{\relax}
\BIBdecl

\bibitem{goodfellow2014generative} 
I.~Goodfellow, \emph{et al.}, ``Generative adversarial nets,'' in \emph{Proc. NIPS 2014} (Montreal, Quebec, Canada), Dec.~8-13, 2014, pp.~1--9.

\bibitem{Thies_2016_CVPR} 
J.~Thies, \emph{et al.}, ``Face2Face: Real-time face capture and reenactment of RGB videos,'' in \emph{Proc. CVPR 2016} (Las Vegas, NV, USA), Jun.~26-Jul.~1, 2016, pp.~2387--2395.

\bibitem{suwajanakorn2017synthesizing} 
S.~Suwajanakorn, S.~M.~Seitz, and I.~Kemelmacher-Shlizerman, ``Synthesizing Obama: Learning lip sync from audio,'' \emph{ACM Trans. Graphics}, vol.~36, no.~4, article\,no.~95, pp.~1--13, 2017.

\bibitem{kim2018deep} 
H.~Kim, \emph{et al.}, ``Deep video portraits,'' \emph{ACM Trans. Graphics}, vol.~37, no.~4, article\,no.~163, pp.~1--14, 2018.

\bibitem{karras2017progressive} 
T.~Karras, T.~Aila, S.~Laine, and J.~Lehtinen, ``Progressive growing of GANs for improved quality, stability, and variation,'' in \emph{Proc. ICLR 2018} (Vancouver, BC, Canada), Apr.~30-May~3, 2018, pp.~1--26.

\bibitem{karras2018style} 
T.~Karras, S.~Laine, and T.~Aila, ``A style-based generator architecture for generative adversarial networks,'' in \emph{Proc. CVPR 2019} (Long Beach, CA, USA), Jun.~16-20, 2019, pp.~4401--4410.

\bibitem{survey_chesney_citron_2018} 
R.~Chesney and D.~K.~Citron, ``Deep Fakes: A looming challenge for privacy, democracy, and national security,'' \emph{California Law Review}, vol.~107, no.~ 6, pp.~1753--1820, 2019.

\bibitem{li2018ictu} 
Y.~Li, M.-C.~Chang, and S.~Lyu, ``In Ictu Oculi: Exposing AI generated fake face videos by detecting eye blinking,'' in \emph{Proc. WIFS 2018} (Hong Kong, China), Dec.~11-13, 2018, pp.~1--7.

\bibitem{afchar2018mesonet} 
D.~Afchar, V.~Nozick, J.~Yamagishi, and I.~Echizen, ``MesoNet: A compact facial video forgery detection network,''  in \emph{Proc. WIFS 2018} (Hong Kong, China), Dec.~11-13, 2018, pp.~1--7.

\bibitem{li2021frequencyaware} 
J.~Li, \emph{et al.}, ``Frequency-aware discriminative feature learning supervised by single-center loss for face forgery detection,'' in \emph{Proc. CVPR 2021}, Jun.~19-25, 2021, pp.~6458--6467.

\bibitem{sun2021improving} 
Z.~Sun, \emph{et al.}, ``Improving the efficiency and robustness of deepfakes detection through precise geometric features,'' in \emph{Proc. CVPR 2021}, Jun.~19-25, 2021, pp.~3609--3618.

\bibitem{he2021forgerynet} 
Y.~He, \emph{et al.},  ``ForgeryNet: A versatile benchmark for comprehensive forgery analysis,'' in \emph{Proc. CVPR 2021}, Jun.~19-25, 2021, pp.~ 4358--4367.

\bibitem{rossler2019faceforensics} 
A.~R{\" o}ssler, \emph{et al.}, ``FaceForensics++: Learning to detect manipulated facial images,'' in \emph{Proc. ICCV 2019} (Seoul, South Korea), Oct.~27-Nov.~2, 2019, pp.~1--11.

\bibitem{li2020celeb} 
Y.~Li, \emph{et al.}, ``Celeb-DF: A large-scale challenging dataset for deepfake forensics,'' in \emph{Proc. CVPR 2020}, Jun.~14-19, 2020, pp.~3207--3216.

\bibitem{li2019exposing} 
Y.~Li and S.~Lyu, ``Exposing deepfake videos by detecting face warping artifacts,'' in \emph{Proc. CVPRW, 2019} (Long Beach, CA, USA),  Jun.~16-17, 2019. pp.~46--52.

\bibitem{Li-etal2020learning} 
L.~Li, \emph{et al.}, ``Face X-ray for more general face forgery detection,'' in \emph{Proc. CVPR 2020}, Jun.~14-19, 2020, pp.~5001--5010.

\bibitem{zhao2021learning} 
T.~Zhao, \emph{et al.}, ``Learning self-consistency for deepfake detection,'' in \emph{Proc. ICCV 2021}, Oct.~11-17, 2021, pp.~15023--15033.

\bibitem{shiohara2022detecting} 
K.~Shiohara and T.~Yamasaki, ``Detecting deepfakes with self-blended images,'' in \emph{Proc. CVPR 2022} (New Orleans, LA, USA), Jun.~19-24, 2022, pp.~18720--18729.

\bibitem{qian2020thinking} 
Y.~Qian, \emph{et al.}, ``Thinking in frequency: Face forgery detection by mining frequency-aware clues,'' in \emph{Proc. ECCV 2020} (Glasgow, UK), Aug.~23-28, 2020, pp.~1--18.

\bibitem{liu2021spatial} 
H.~Liu, \emph{et al.}, ``Spatial-phase shallow learning: Rethinking face forgery detection in frequency domain,'' in \emph{Proc. CVPR 2021}, Jun.~19-25, 2021, pp.~772--781.

\bibitem{luo2021generalizing} 
Y.~Luo, Y.~Zhang, J.~Yan, and W.~Liu, ``Generalizing face forgery detection with high-frequency features,'' in \emph{Proc. CVPR 2021}, Jun.~19-25, 2021, pp.~16317--16326.

\bibitem{dong2022think} 
C.~Dong, A.~Kumar, and E.~Liu, ``Think twice before detecting GAN-generated fake images from their spectral domain imprints,'' in \emph{Proc. CVPR 2022} (New Orleans, LA, USA), Jun.~18-24, 2022, pp.~7855--7864.

\bibitem{cozzolino2018forensictransfer} 
D.~Cozzolino, \emph{et al.}, ``ForensicTransfer: Weakly-supervised domain adaptation for forgery detection,'' \emph{arXiv:1812.02510}, 2018.

\bibitem{aneja2020generalized} 
S.~Aneja and M.~Nie{\ss}ner, ``Generalized zero and few-shot transfer for facial forgery detection,'' \emph{arXiv:2006.11863}, 2020.

\bibitem{qiu2022few} 
H.~Qiu, \emph{et al.}, ``Few-shot forgery detection via guided adversarial interpolation,'' \emph{Pattern Recognition}, vol.~144, article\,no.~109863, pp.~1--11, Dec. 2023.

\bibitem{wang2022rethinking} 
H.~Wang, X.~Guo, Z.-H.~Deng, and Y.~Lu, ``Rethinking minimal sufficient representation in contrastive learning,'' in \emph{Proc. CVPR 2022} (New Orleans, LA, USA), Jun.~18-24, 2022, pp.~16041--16050.

\bibitem{chen2020big} 
T.~Chen, \emph{et al.}, ``Big self-supervised models are strong semi-supervised learners,'' in \emph{Proc. NeurIPS 2020}, Dec.~6-12, 2020, pp.~1--13.

\bibitem{tifs_distillation} 
Z.~Li, \emph{et al.}, ``One-class knowledge distillation for face presentation attack detection,'' \emph{IEEE Trans. Information Forensics and Security}, vol.~17, pp.~2137--2150, Jun. 2022.

\bibitem{dosovitskiy2021an} 
A.~Dosovitskiy, \emph{et al.}, ``An image is worth 16x16 words: Transformers for image recognition at scale,'' in \emph{Proc. ICLR 2021}, May~3-7, 2021, pp.~1--21.

\bibitem{resnet_CVPR} 
K.~He, X.~Zhang, S.~Ren, and J.~Sun, ``Deep residual learning for image recognition,'' in \emph{Proc. CVPR 2016} (Las Vegas, NV, USA), Jun.~26-Jul.~1, 2016, pp.~770--778.

\bibitem{chollet2017xception} 
F.~Chollet, ``Xception: Deep learning with depthwise separable convolutions,'' in \emph{Proc. CVPR 2017}  (Honolulu, HI, USA), Jul.~21-26, 2017, pp.~1251--1258.

\bibitem{tan2019efficientnet} 
M.~Tan and Q.~Le, ``EfficientNet: Rethinking model scaling for convolutional neural networks,'' in \emph{Proc. ICML 2019} (Long Beach, CA, USA), Jun.~10-15, 2019, pp.~1--11.

\bibitem{agarwal2019protecting} 
S.~Agarwal, \emph{et al.}, ``Protecting world leaders against deep fakes,'' in \emph{Proc. CVPR 2019 Workshops} (Long Beach, CA, USA), Jun.~16-20, 2019, pp.~38--45.

\bibitem{sun2021domain} 
K.~Sun, \emph{et al.},  ``Domain general face forgery detection by learning to weight,'' in \emph{Proc. AAAI 2021}, Feb.~2-9, 2021, pp.~2638--2646.

\bibitem{chen2022selfsupervised} 
L.~Chen, \emph{et al.}, ``Self-supervised learning of adversarial example: Towards good generalizations for deepfake detection,'' in \emph{Proc. CVPR 2022} (New Orleans, LA, USA), Jun.~19-24, 2022, pp.~18710--18719.

\bibitem{TIFSnew1} 
J.~Yang, \emph{et al.}, ``MTD-Net: Learning to detect deepfakes images by multi-scale texture difference,'' \emph{IEEE Trans. Information Forensics and Security}, vol.~16, pp.~4234--4245, Aug. 2021.

\bibitem{TIFSnew2} 
P.~Yu, \emph{et al.}, ``Improving generalization by commonality learning in face forgery detection,'' \emph{IEEE Trans. Information Forensics and Security}, vol.~17, pp.~547--558, Feb. 2022.

\bibitem{TIFSnew3} 
J.~Wang, Y.~Sun, and J.~Tang, ``LiSiam: Localization invariance Siamese network for deepfake detection,'' \emph{IEEE Trans. Information Forensics and Security}, vol.~17, pp.~2425--2436, Jul. 2022.

\bibitem{TIFSnew4} 
C.~Miao, \emph{et al.}, ``Hierarchical frequency-assisted interactive networks for face manipulation detection,'' \emph{IEEE Trans. Information Forensics and Security}, vol.~17, pp.~3008--3021, Aug. 2022.

\bibitem{TIFSnew5} 
C.~Miao, \emph{et al.}, ``F$^{\rm 2}$Trans: High-frequency fine-grained transformer for face forgery detection,'' \emph{IEEE Trans. Information Forensics and Security}, vol.~18, pp.~1039--1051, Jan. 2023.

\bibitem{zhu2021face} 
X.~Zhu, \emph{et al.}, ``Face forgery detection by 3D decomposition,'' in \emph{Proc. CVPR 2021}, Jun.~19-25, 2021, pp.~2929--2939.

\bibitem{ben2006analysis} 
S.~Ben-David, J.~Blitzer, K.~Crammer, and F.~Pereira, ``Analysis of representations for domain adaptation,'' in \emph{Proc. NIPS 2006} (Vancouver, BC, Canada), Dec.~4-9, 2006, pp.~1--8.

\bibitem{DeepDomainConfusion_CDA_5} 
E.~Tzeng, \emph{et al.}, ``Deep domain confusion: Maximizing for domain invariance,'' \emph{arXiv:1412.3474}, 2014.

\bibitem{long2015learning} 
M.~Long, Y.~Cao, J.~Wang, and M.~Jordan, ``Learning transferable features with deep adaptation networks,'' in \emph{Proc. ICML 2015} (Lille, France), Jul.~6-11, 2015. pp.~97--105.

\bibitem{DeepCORAL_ECCV_CDA_4} 
B.~Sun and K.~Saenko, ``Deep CORAL: Correlation alignment for deep domain adaptation,'' in \emph{Proc ECCV 2016 Workshops} (Amsterdam, The Netherlands), Oct.~8-10 and Oct.~15-16, 2016, pp.443--450.

\bibitem{zhang2019bridging_mdd} 
Y.~Zhang, T.~Liu, M.~Long, and M.~I. Jordan, ``Bridging theory and algorithm for domain adaptation,'' in \emph{Proc. ICML 2019} (Long Beach, CA, USA), Jun.~9-15, 2019, pp.~7404--7413.

\bibitem{goodfellow2020generative} 
I.~Goodfellow, \emph{et al.}, ``Generative adversarial networks,'' \emph{Communications of the ACM}, vol.~63, no.~11, pp.~139--144, 2020.

\bibitem{DANN_JMLR_ODA_5} 
Y.~Ganin, \emph{et al.}, ``Domain-adversarial training of neural networks,'' \emph{J. Machine Learning Research}, vol.~17, no.~1, pp.~2096--2030, 2016.

\bibitem{tzeng2017adversarial} 
E.~Tzeng, J.~Hoffman, K.~Saenko, and T.~Darrell, ``Adversarial discriminative domain adaptation,'' in \emph{Proc. CVPR 2017} (Honolulu, HI, USA), Jul.~21-26, 2017, pp.~7167--7176.

\bibitem{xu2020adversarial} 
M.~Xu, \emph{et al.}, ``Adversarial domain adaptation with domain mixup,'' in \emph{Proc. AAAI 2020} (New York, NY, USA), Feb.~7-12, 2020, pp.~6502--6509.

\bibitem{yang2023tvt} 
J.~Yang, J.~Liu, N.~Xu, and J.~Huang, ``TVT: Transferable vision transformer for unsupervised domain adaptation,'' in \emph{Proc. WACV 2023} (Waikoloa, HI, USA), Jan.~3-7, 2023, pp.~520--530.

\bibitem{xu2021cdtrans} 
T.~Xu, \emph{et al.},  ``CDTrans: Cross-domain transformer for unsupervised domain adaptation,'' \emph{arXiv:2109.06165}, 2021.

\bibitem{sun2022safe_ssrt} 
T.~Sun, C.~Lu, T.~Zhang, and H.~Ling, ``Safe self-refinement for transformer-based domain adaptation,'' in \emph{Proc. CVPR 2022} (New Orleans, LA, USA), Jun.~19-24, 2022, pp.~7191--7200.

\bibitem{chen2021featuretransfer} 
B.~Chen and S.~Tan, ``FeatureTransfer: Unsupervised domain adaptation for cross-domain deepfake detection,'' \emph{Security and Communication Networks}, vol.~2021, article~9942754, pp.~1--8, 2021.

\bibitem{li2019exploring} 
Y.~Li, X.~Bian, M.-C.~Chang, and S.~Lyu, ``Exploring the vulnerability of single shot module in object detectors via imperceptible background patches,'' in \emph{Proc. BMVC 2019} (Cardiff, UK), Sep.~9-12, 2019, pp.~1--12.

\bibitem{selvaraju2017grad} 
R.~R.~Selvaraju, \emph{et al.}, ``Grad-CAM: Visual explanations from deep networks via gradient-based localization,'' in \emph{Proc. ICCV 2017} (Venice, Italy), Oct.~22-29, 2017, pp.~618--626.

\bibitem{vaswani2017attention} 
A.~Vaswani, \emph{et al.}, ``Attention is all you need,'' \emph{Proc. NIPS 2017} (Long Beach, CA, USA), Dec.~4-9, 2017, pp.~1--11.

\bibitem{stylegan} 
T.~Karras, S.~Laine, and T.~Aila, ``A style-based generator architecture for generative adversarial networks,'' \emph{IEEE Trans. Pattern Analysis and Machine Intelligence}, vol.~43, no.~12, pp.~4217--4228, Dec. 2021.

\bibitem{Deng2020CVPR}. 
J.~Deng, \emph{et al.}, ``RetinaFace: Single-shot multi-level face localisation in the wild,'' in \emph{Proc. CVPR 2020} (Seattle, WA, USA), Jun.~13-19, 2020, pp.~5202--5211.

\bibitem{pytoch_ARXIV} 
A.~Paszke, \emph{et al.}, ``Automatic differentiation in pytorch,'' in \emph{Proc. NIPS 2017} (Long Beach, CA, USA), Dec.~4-9, 2017, pp.~1--4.

\bibitem{zhao2021multiattentional} 
H.~Zhao, \emph{et al.}, ``Multi-attentional deepfake detection,'' in \emph{Proc. CVPR 2021}, Jun.~19-25, 2021, pp.~2185--2194.

\bibitem{cao2022endtoend} 
J.~Cao, \emph{et al.}, ``End-to-end reconstruction-classification learning for face forgery detection,'' in \emph{Proc. CVPR 2022} (New Orleans, LA, USA), Jun.~18-24, 2022, pp.~4113--4122.
	
\bibitem{fei2022learning} 
J.~Fei, \emph{et al.}, ``Learning second order local anomaly for general face forgery detection,'' in \emph{Proc. CVPR 2022} (New Orleans, LA, USA), Jun.~18-24, 2022, pp.~20270--2080.

\bibitem{zheng2021exploring} 
Y.~Zheng, \emph{et al.}, ``Exploring temporal coherence for more general video face forgery detection,'' in \emph{Proc. ICCV 2021}, Oct.~11-17, 2021, pp.~15044--15054.

\bibitem{van2008visualizing} 
L.~Van~der Maaten and G.~Hinton, ``Visualizing data using t-SNE.'' \emph{J. Machine Learning Research}, vol.~9, article\,86, pp.~2579--2605, 2008.

\bibitem{twobranch} 
I.~Masi, \emph{et al.}, ``Two-branch recurrent network for isolating deepfakes in videos,'' in \emph{Proc. ECCV 2020} (Glasgow, UK), Aug.~23-28, 2020, pp.~667--684.

\bibitem{zhou2021faceinwild} 
T.~Zhou, W.~Wang, Z.~Liang, and J.~Shen, ``Face forensics in the wild,'' in \emph{Proc. CVPR 2021}, Jun.~19-25, 2021, pp.~5778--5788.
	

\bibitem{haliassos2021lips} 
A.~Haliassos, K.~Vougioukas, S.~Petridis, and M.~Pantic, ``Lips don't lie: A generalisable and robust approach to face forgery detection,'' in  \emph{Proc. CVPR 2021}, Jun.~19-25, 2021, pp.~5039--5049.

\bibitem{ganin2015unsupervised} 
Y.~Ganin and V.~Lempitsky, ``Unsupervised domain adaptation by backpropagation,'' in \emph{Proc. ICML} (Lille, France), Jul.~6-11, 2015, pp.~1180--1189.

\bibitem{entropy} 
Y.~Grandvalet and Y.~Bengio, ``Semi-supervised learning by entropy minimization,'' in \emph{Proc. NIPS 2004} (Vancouver, BC, Canada), Dec.~1, 2004, pp.~529--536.

\bibitem{dfdcp_dataset}
B.~Dolhansky, R.~Howes, B.~Pflaum, N.~Baram and C.~Canton-Ferrer,``The Deepfake Detection Challenge (DFDC) Preview Dataset,`` {\em ArXiv:1910.08854}, 2019


\bibitem{wang2023altfreezing}Z.~Wang, J.~Bao, W.~Zhou, W.~Wang and H.~Li,``AltFreezing for More General Video Face Forgery Detection,`` {\em Proc. CVPR 2023}. pp. 4129-4138 (2023)

\bibitem{xu2023tall}Y.~Xu, J.~Liang, G.~Jia, Z.~Yang, Y.~Zhang and R.~He,``TALL: Thumbnail Layout for Deepfake Video Detection,`` {\em Proc. ICCV 2023}. pp. 22601-22611 (2023)

\bibitem{wang2023dire}Z.~Wang, J.~Bao, W.~Zhou, W.~Wang, H.~Hu, H.~Chen and H.~Li,``DIRE for Diffusion-Generated Image Detection,`` {\em Proc. ICCV 2023}. pp. 22388-22398 (2023)

\bibitem{lee2013pseudolabel}D.~Lee,``Pseudo-Label : The Simple and Efficient Semi-Supervised Learning Method for Deep Neural Networks,`` {\em Proc. ICML} 2013

\bibitem{yalniz2019billionscale}I.~Yalniz,``Billion-scale semi-supervised learning for image classification,`` {\em ArXiv:1905.00546}, 2019

\bibitem{tarvainen2017meanteacher}A.~Tarvainen, and H.~Valpola,``Mean teachers are better role models: Weight-averaged consistency targets improve semi-supervised deep learning results``, {\em Proc. ICLR}. pp. 1195-1204 2017


\end{thebibliography}


\begin{IEEEbiography}[{\includegraphics[width=1in,height=1.25in,clip,keepaspectratio]{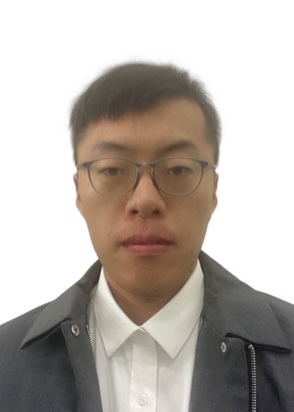}}]{Qingxuan Lv}
was born in Shanxi, China, in 1996. He received his bachelor's degree in Computer Science and Technology from the Shanxi University of Finance and Economics in 2018. He received his master's degree in Computer Science and Technology from the Ocean University of China (OUC) in 2021. He is currently a candidate of a doctor's degree at the ocean group of VisionLab OUC. His research interests include computer vision and machine learning. Specifically, he is interested in universal domain adaptation and semantic segmentation.
\end{IEEEbiography}


\begin{IEEEbiography}[{\includegraphics[width=1in,height=1.25in,clip,keepaspectratio]{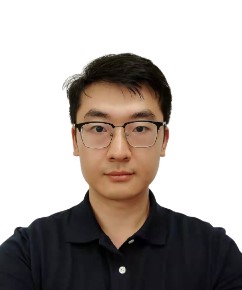}}]{Yuezun Li} (Member, IEEE) received the B.S. degree in software engineering from Shandong University in 2012, the M.S. degree in computer science in2015, and the Ph.D. degree in computer science from University at Albany–SUNY, in 2020. He was a Senior Research Scientist with the Department of Computer Science and Engineering, University at Buffalo–SUNY. He is currently a Lecturer with the Center on Artifcial Intelligence, Ocean University of China. His research interests include computer vision and multimedia forensics. His work has been published in peer reviewed conferences and journals, including ICCV, CVPR, TIFS, TCSVT, etc. 
\end{IEEEbiography}


\begin{IEEEbiography}[{\includegraphics[width=1in,height=1.25in,clip,keepaspectratio]{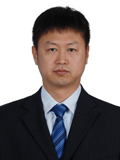}}]{Junyu Dong} received the B.Sc. and M.Sc. degrees in applied mathematics from the Department of Applied Mathematics, Ocean University of China, Qingdao, China, in 1993 and 1999, respectively, and the Ph.D. degree in image processing from the Department of Computer Science, Heriot-Watt University, Edinburgh, U.K., in November 2003. He is currently a Professor and the Head of the Department of Computer Science and Technology. His research interests include machine learning, big data, computer vision, and underwater image processing.
\end{IEEEbiography}

\begin{IEEEbiography}[{\includegraphics[width=0.9in,height=1.1in,clip,keepaspectratio]{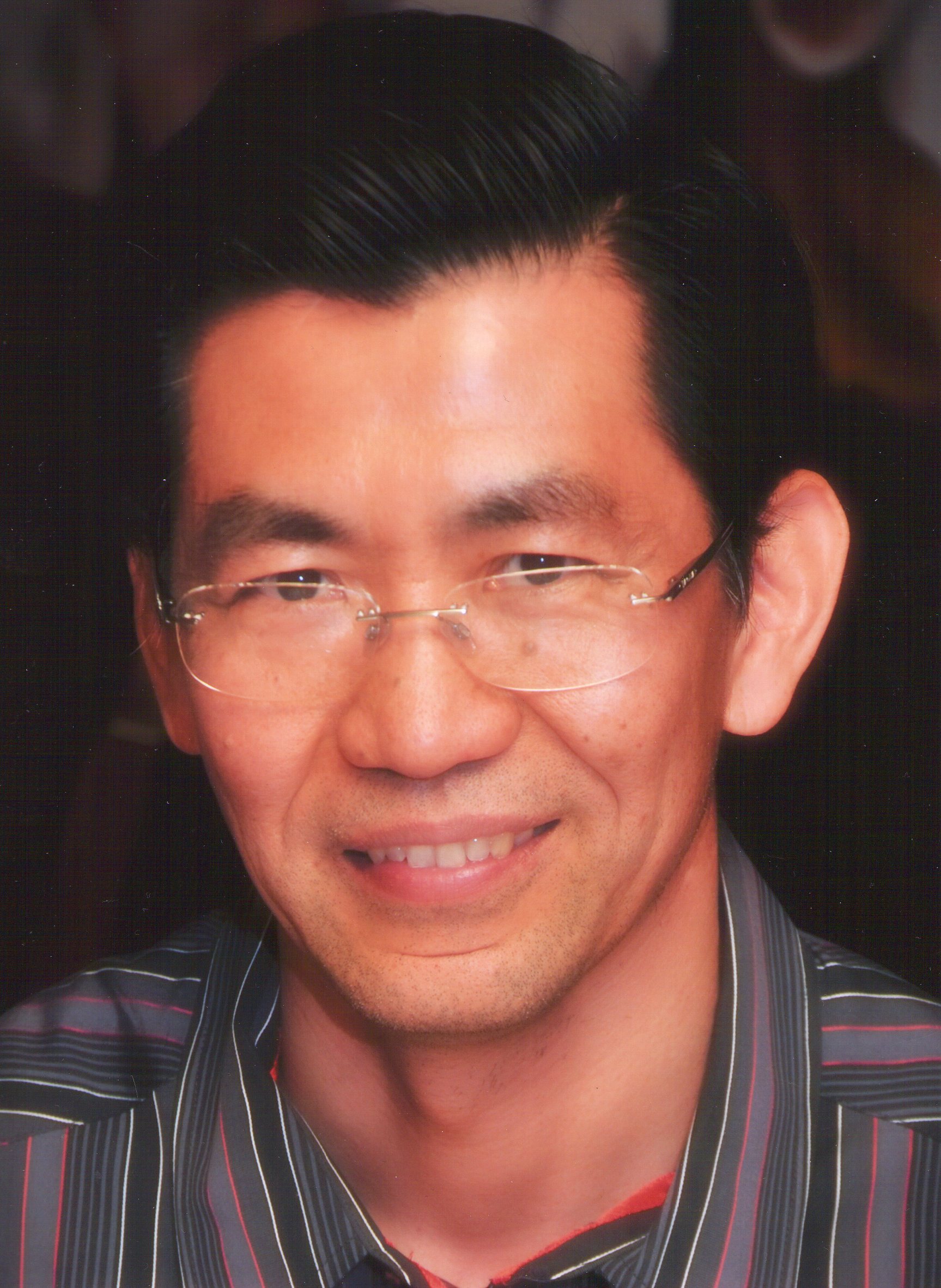}}]{Sheng Chen} (IEEE Life Fellow) received the B.Eng. degree in control engineering from the East China Petroleum Institute, Dongying, China, in 1982, the Ph.D. degree in control engineering from City University, London, in 1986, and the higher doctoral (D.Sc.) degree from the University  of Southampton, Southampton, U.K., in 2005. From 1986 to 1999, he held research and academic appointments with the University of Sheffield, the University of Edinburgh, and the University of Portsmouth, U.K. Since 1999, he has been with the School of Electronics and Computer Science, University of Southampton, where he is a Professor of Intelligent Systems and Signal Processing. His research interests include adaptive signal processing, wireless communications, modeling and identification of nonlinear systems, neural network and machine learning, intelligent control system design, evolutionary computation methods, and optimization. Professor Chen has authored over 700 research papers. He have 20,000+ Web of Science citations with h-index 61, and 39,000+ Google Scholar citations with h-index 83. Dr Chen was elected to a fellow of the United Kingdom Royal Academy of Engineering in 2014. He is a fellow of Asia-Pacific Artificial Intelligence Association (FAAIA), a fellow of IET, and an original ISI Highly Cited Researcher in engineering (March 2004).

\end{IEEEbiography}

\begin{IEEEbiography}[{\includegraphics[width=1in,height=1.25in,clip,keepaspectratio]{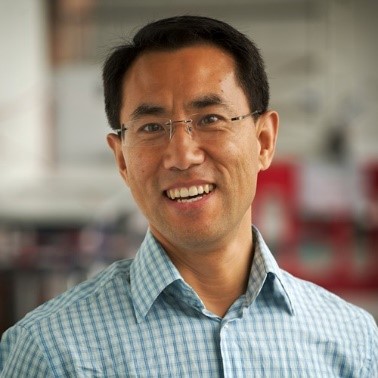}}]{Hui Yu} 
is a Professor of Visual Computing with the University of Portsmouth, UK and leads the Visual Computing Group. He received the PhD degree from Brunel University London in 2009. He worked at the University of Glasgow and Queen's University Belfast before joining the University of Portsmouth in 2012. His research interests include visual computing and machine learning with application to 3D/4D facial expression tracking, reconstruction and affective analysis, human-machine and social interaction, virtual and augmented reality as well as intelligent vehicles. He serves as an Associate Editor for IEEE Transactions on Human-Machine Systems, IEEE Transactions on Intelligent Vehicles, and IEEE Transactions on Computational Social Systems journal.  
\end{IEEEbiography}

\begin{IEEEbiography}[{\includegraphics[width=1in,height=1.25in,clip,keepaspectratio]{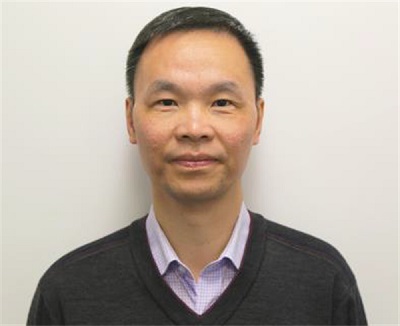}}]{Huiyu Zhou} received the B.Eng. degree in radio technology from the Huazhong University of Science and Technology, Wuhan, China, in 1990, the M.Sc. degree in biomedical engineering from the University of Dundee, Dundee, U.K., in 2002, and the Ph.D. degree in computer vision from Heriot-Watt University, Edinburgh, U.K., in 2006. He is currently a Full Professor with the School of Computing and Mathematical Sciences, University of Leicester, Leicester, U.K. He has authored or coauthored over 450 peer-reviewed papers in the field.
\end{IEEEbiography}

\begin{IEEEbiography}[{\includegraphics[width=1in,height=1.25in,clip,keepaspectratio]{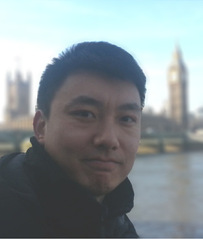}}]{Shu Zhang} 
Shu Zhang is currently an Associate Professor and Postgraduate supervisor at Ocean University of China, Qingdao, China. He received his PhD in Computer Application Technologies from Ocean University of China. He was previously a research associate at the University of Portsmouth, Portsmouth, UK. His main research interests include computer vision, feature analysis, 3D reconstruction, video processing, underwater image analysis, and deep learning among others. 
\end{IEEEbiography}

\end{document}